\documentclass[lettersize,journal]{IEEEtran}
\usepackage{amsmath,amsfonts}
\usepackage{algorithmic}
\usepackage{algorithm}
\usepackage{array}
\usepackage[caption=false,font=normalsize,labelfont=sf,textfont=sf]{subfig}
\usepackage{textcomp}
\usepackage{stfloats}
\usepackage{url}
\usepackage{verbatim}
\usepackage{graphicx}
\usepackage{cite}
\usepackage{color}
\usepackage{makecell}

\usepackage{cleveref}

\usepackage{graphicx}
\usepackage{amsmath}
\usepackage{amssymb}
\usepackage{booktabs}
\usepackage{multirow}
\usepackage{enumitem}
\usepackage{multirow}
\usepackage{float}

\usepackage{diagbox}

\usepackage{algorithmic}
\usepackage{algorithm}

\usepackage{caption}
\usepackage{cleveref}

\renewcommand{\algorithmicrequire}{\textbf{Input:}}
\renewcommand{\algorithmicensure}{\textbf{Output:}}

\def\eg{\emph{e.g.}}

\def\etc{\emph{etc}} \def\vs{\emph{vs.}}
 
\def\etal{\emph{et al.}}

\usepackage{colortbl}
\usepackage{xcolor}

\definecolor{color1}{RGB}{237, 237, 237}

\begin{document}

\title{PSDiff: Diffusion Model for Person Search with Iterative and Collaborative Refinement}

\author{Chengyou Jia,
        Minnan Luo$^*$,
        Zhuohang Dang, 
        Guang Dai,
        Xiaojun Chang,
        and~Jingdong Wang
        \thanks{Copyright © 2024 IEEE. Personal use of this material is permitted. However, permission to use this material for any other purposes must be obtained from the IEEE by sending an email to pubs-permissions@ieee.org.}
	\thanks{$^*$Corresponding author: Minnan Luo.}
        \thanks{This work is supported by the National Nature Science Foundation of China (No. 62192781, No. 62272374), the Natural Science Foundation of Shaanxi Province (No. 2024JC-JCQN-62), the National Nature Science Foundation of China (No. 62250009, No. 62137002), Project of China Knowledge Center for Engineering Science and Technology, Project of Chinese academy of engineering ``The Online and Offline Mixed Educational Service System for ‘The Belt and Road’ Training in MOOC China'', and the K. C. Wong Education Foundation.}
	\thanks{Chengyou Jia, Minnan Luo, and Zhuohang Dang are with the School of Computer Science and Technology, Xi'an Jiaotong University, and Key Laboratory of Intelligent Networks and Network Security (Xi'an Jiaotong University), Ministry of Education, Xi'an, Shaanxi 710049, China, {e-mail: \{cp3jia,dangzhuohang\}@stu.xjtu.edu.cn,\\ \{qhzheng,minnluo\}@mail.xjtu.edu.cn}.}
	\thanks{Guang Dai is with SGIT AI Lab, and also with State Grid Shaanxi Electric Power Company Limited, State Grid Corporation of China, Shaanxi, China, {e-mail: guang.gdai@gmail.com}.}
	\thanks{Xiaojun Chang is with the School of Information Science and Technology, University of Science and Technology of China. Xiaojun Chang is also a Visiting Professor with Department of Computer Vision, Mohamed bin Zayed University of Artificial Intelligence (MBZUAI). {e-mail: cxj273@gmail.com.}}
        \thanks{Jingdong Wang is with the Baidu Inc, China, {e-mail: wangjingdong@outlook.com.}}
}

\markboth{Journal of \LaTeX\ Class Files,~Vol.~14, No.~8, August~2021}%
{Shell \MakeLowercase{\textit{et al.}}: A Sample Article Using IEEEtran.cls for IEEE Journals}


\maketitle

\begin{abstract}
Dominant Person Search methods aim to localize and recognize query persons in a unified network,  which jointly optimizes the two sub-tasks of pedestrian detection and Re-Identification (ReID). Despite significant progress, current methods face two primary challenges: 1) the pedestrian candidates learned within detectors are suboptimal for the ReID task. 2) the potential for collaboration between two sub-tasks is overlooked.
To address these issues, we present a novel \textbf{P}erson \textbf{S}earch framework based on the \textbf{Diff}usion model, \textbf{PSDiff}. PSDiff formulates the person search as a dual denoising process from noisy boxes and ReID embeddings to ground truths. Distinct from the conventional Detection-to-ReID approach, our denoising paradigm discards prior pedestrian candidates generated by detectors, thereby avoiding the local optimum problem of the ReID task. Following the new paradigm, we further design a new Collaborative Denoising Layer (\textbf{CDL}) to optimize detection and ReID sub-tasks in an iterative and collaborative way, which makes two sub-tasks mutually beneficial. Extensive experiments on the standard benchmarks show that PSDiff achieves state-of-the-art performance with fewer parameters and elastic computing overhead.
\end{abstract}

\section{Introduction}
\IEEEPARstart{P}{erson} Search \cite{xiaoli2017joint, Zheng_2017_CVPR} aims to locate and recognize query persons in a gallery of unconstrained scene images, which consists of two sub-tasks, detection and Re-IDentification (ReID). 
Because of acting on the whole scene images, person search shows more potential than the ReID task \cite{7918589,8693885,8485427,7864367} in real-world applications, such as criminal tracking, person behavior analysis and human-computer interaction. 
Early approaches \cite{Zheng_2017_CVPR, 9003518} adopt the two-stage paradigm that tackles two sub-tasks separately. In detail, a detector is applied to scene images to predict the bounding boxes and then a followed ReID network extracts ReID embeddings from the detected person slices. In contrast, end-to-end methods  \cite{xiaoli2017joint,yan2019learning, chen2020norm, li2021sequential, DBLP:conf/cvpr/YanLQBL00021,dong2020instance, Cao_PSTR_CVPR_2022,yu2022coat} aim to localize and recognize query persons in a unified network, which is more concise and efficient in general. Due to their advantages, end-to-end methods became the dominant fashion and attracted the attention of researchers.

\setlength{\belowcaptionskip}{-0.0cm}
\begin{figure}[t]
  \centering
     \centering
     \includegraphics[width=0.47\textwidth]{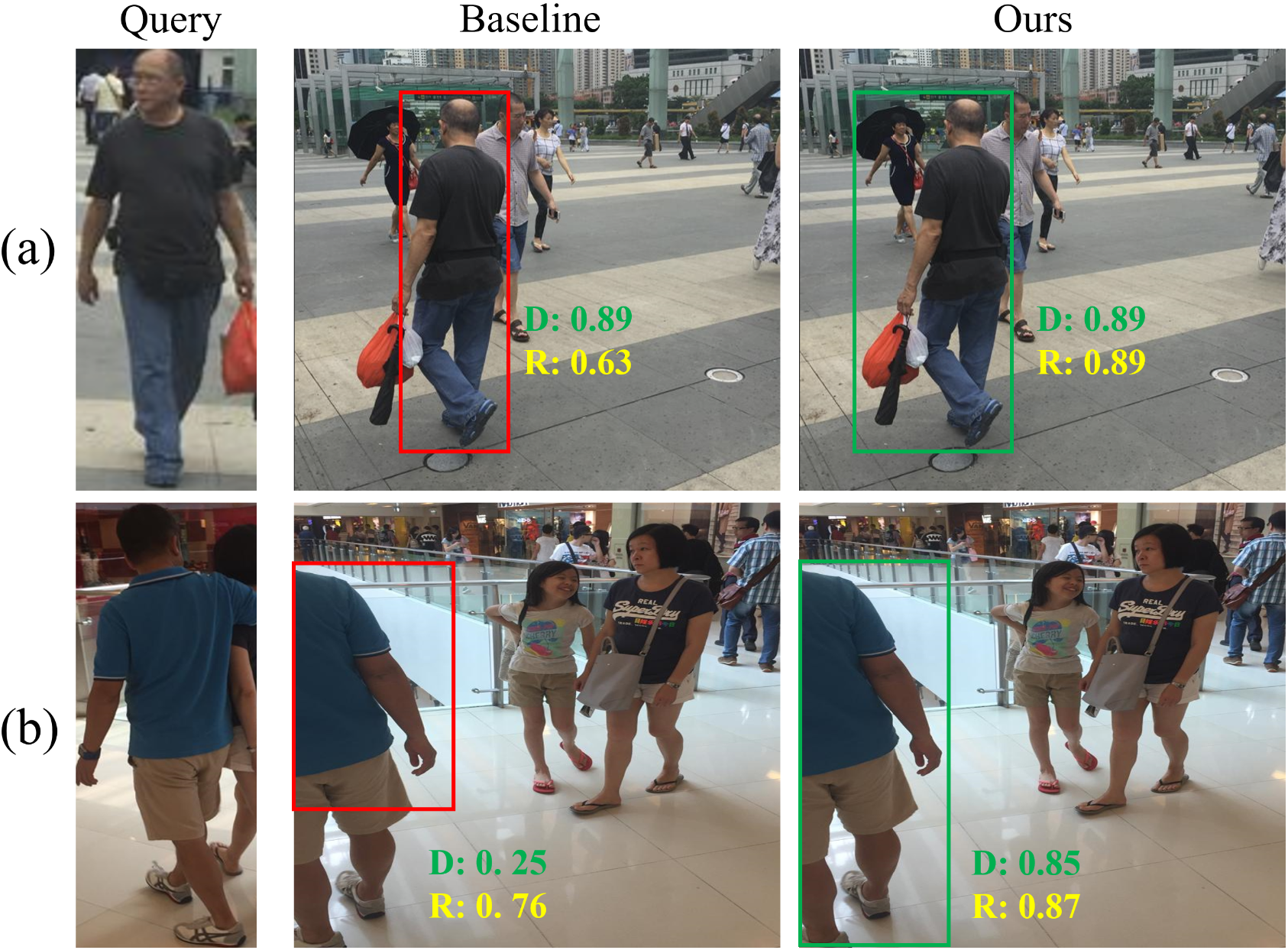}
     \label{fig:intro}
   \caption{(a) The baseline with prior pedestrian candidates ignores some crucial parts such as carry-on items, causing low ReID matching scores. (b) The baseline without collaboration misses results with high ReID matching scores due to inaccurate positions and low confidence scores. In comparison, our method can increase discriminative parts and refine two tasks collaboratively, producing more discriminative embeddings and more accurate detection results. ``D'' and ``R'' index confidence scores of detection and ReID matching scores with queries, respectively. The bounding boxes in green and red denote the correct and wrong results.}
   \label{fig:intro}
\end{figure}

\setlength{\belowcaptionskip}{-0.0cm}

Although previous end-to-end methods have made significant progress, two major challenges still remain. (1) \textbf{\textit{the pedestrian candidates learned within detectors are suboptimal for the ReID task.}} Existing end-to-end methods \cite{xiaoli2017joint,chen2020norm,li2021sequential,dong2020instance,Cao_PSTR_CVPR_2022,yu2022coat} are first built on traditional detectors, like Faster-RCNN \cite{rennips15fasterrcnn} or DETR \cite{carion2020end}, which inevitably bring empirical or learnable pedestrian candidates to the person search framework. Specifically, Faster-RCNN introduces the Region Proposal Network (RPN) to generate high-quality region proposals and DETR proposes learnable object queries to learn potential positions. The purpose of these modules is to learn proposals of pedestrians in advance according to given annotations. As shown in Fig. \ref{fig:intro} (a), some crucial parts, which possess the discriminative information, are essential for ReID matching but may be ignored, leading to unsatisfactory ReID matching scores. (2) \textbf{\textit{The potential for collaboration between two sub-tasks is overlooked.}} Existing methods \cite{xiaoli2017joint,chen2020norm,li2021sequential} only follow the principle that more accurate detection results contribute more to the learning of ReID tasks but ignore that better ReID clues can also bring more high-quality detection results. Fig. \ref{fig:intro} (b) shows detection candidates with low confidences and inaccurate positions can be effectively improved by high ReID matching scores. Therefore, two sub-tasks should collaboratively work to benefit each other. 

Inspired by \cite{ho2020denoising, Brempong_2022_CVPR, chen2022diffusiondet}, we address the aforementioned issues by designing a novel \textbf{P}erson \textbf{S}earch framework based on the \textbf{Diff}usion model, called \textbf{PSDiff}. Innovatively, we formulate the person search task as a dual denoising process from noisy boxes and ReID embeddings to ground truths. Specifically, during the training stage, both object boxes and ReID embeddings diffuse from ground truths to randomly distributed noises by a designed dual noise generator. In this manner, PSDiff effectively sidesteps the reliance on prior pedestrian candidates, thereby equitably aligning the two tasks and avoiding the pitfall of sub-optimization focused on a singular aspect.
Then, the model learns a novel Collaborative Denoising Layer (CDL) to reverse the noising process. The proposed CDL generates collaborative features by interacting features from two sub-tasks and then predicts results with a cascaded architecture. 
During the inference stage, the trained CDL iteratively and collaboratively refines a set of randomly generated boxes and ReID embeddings.
This design enables our denoising paradigm to efficiently facilitate a collaborative interplay between the two sub-tasks, thereby ensuring their mutual enhancement.

We evaluate the PSDiff on CUHK-SYSU and PRW datasets. With ResNet-50 and Swin-Base, PSDiff achieves 95.1\% and 95.7\% mAP on CUHK-SYSU, and 53.5\% and 57.1\% mAP on PRW, outperforming existing methods. Besides, our approach also shows several appealing properties with fewer parameters and elastic computing overhead. 

Our main contributions are summarized as follows: 
\begin{itemize}
\item We propose \textbf{PSDiff}, a novel person search framework based on the diffusion model. By formulating the person search task as a dual denoising process, PSDiff successfully addresses
the challenges of destructive pedestrian candidates and missing collaboration. To our knowledge, PSDiff is the first study to apply the diffusion model to person search. 
\item We propose the Collaborative Denoising Layer (CDL), which aims to denoise noisy boxes and ReID embeddings. Following the denoising paradigm, the trained CDL can be reused to refine predictions in an iterative and collaborative way.
\item The extensive experiments on the standard benchmarks, including CUHK-SYSU
and PRW, show that PSDiff achieves state-of-the-art performance. What's more, PSDiff also shows several appealing properties about parameters and computing overhead.
\end{itemize}

The remainder of the article is organized as follows. In Section \ref{sec:realted}, we briefly summarize the related works, \emph{e.g.}, person search and diffusion model. Section \ref{sec:preliminaries} introduces the formulation of diffusion models. In
Section \ref{sec:method}, we describe the proposed PSDiff in detail. In Sections \ref{sec:exp}, we first present datasets and implementation details and then report experimental results comparing with SOTA methods. We also conduct the ablation study and provide more detailed analysis. Section \ref{sec:conclusion} concludes this paper and describes some ethical considerations of this work. 

\section{Related Work}
\label{sec:realted}
\subsection{Person Search}

The person search task is built upon the foundation of person Re-IDentification (ReID) that aims at matching a target individual from a gallery of cropped pedestrian images. Advances in the field of ReID not only led to key technical improvements, such as metric learning \cite{ye2021deep}, but also introduced practical settings like Unsupervised ReID \cite{wang2020unsupervised,9969623,han2021group}, Domain Adaptation ReID \cite{ge2020self,zhao2020unsupervised,gu2023color}, and Video-based ReID \cite{mclaughlin2016recurrent,9416694}. However, a key limitation of person ReID is the lack of manually cropped bounding boxes in real-world scenarios. This led to the emergence of person search, which combines person detection and ReID in unconstrained scenes. Person search has attracted significant attention in the computer vision community due to its real-world applications, such as criminal tracking \cite{9540797}, person behavior analysis \cite{6898845}, and human-computer interaction \cite{4543858,7208833}. Existing solutions to person search generally fall into two 
categories: two-stage methods and end-to-end methods.

\subsubsection{Two-stage methods}  these methods \cite{Zheng_2017_CVPR,han2019re,9003518,wang2020tcts} tackle detection and ReID \cite{8640834,8481710,9285312,6490028,7948787,7979595} tasks separately. They employ the independent detector and ReID network to predict the bounding boxes and extract ReID features, respectively. Zheng \etal \cite{Zheng_2017_CVPR} pioneered a two-stage framework and evaluated the method on various combinations of different detectors \cite{7279146,8328854,9282190} and ReID networks \cite{7918589,8693885,8485427,7864367}. They further proposed the Confidence Weighted Similarity (CWS) to integrate detection confidence into the process of similarity matching. Han \etal \cite{han2019re} proposed a ReID-driven localization refinement framework, which employs the supervision of ReID loss to produce more reliable bounding boxes. Wang \etal \cite{wang2020tcts} further pointed out the consistency problem that the ReID model trained with ground-truth bounding boxes is unreliable and proposed a task-consistent framework by producing query-like bounding boxes to alleviate this issue. 
In general, two-stage methods focus on how to learn more discriminative Re-ID features based on the detection results. Despite the excellent performance, two-stage methods always suffer from terrible efficiency due to the requirement of training and inference with two networks.

\subsubsection{End-to-end methods} Xiao \etal \cite{xiaoli2017joint} first proposed an end-to-end framework by employing the Faster RCNN as the detector and sharing base layers with the ReID network. After that, end-to-end methods have attracted the attention of researchers and have been the dominant fashion \cite{yan2019learning, chen2020norm, li2021sequential, DBLP:conf/cvpr/YanLQBL00021,dong2020instance, Cao_PSTR_CVPR_2022,yu2022coat}. NAE \cite{chen2020norm} proposed to disentangle the person embedding into norm and angle for balancing two conflicting sub-tasks. 
Li \etal \cite{li2021sequential} proposed a Sequential End-to-end Network (SeqNet), which tackles this problem with two sequential sub-networks (detector and ReID module) and refines the predicted bounding boxes of the early stage. Furthermore, Yan \etal \cite{DBLP:conf/cvpr/YanLQBL00021}  first employed the anchor-free framework to avoid the high computational overhead from dense object anchors. They further introduced an aligned feature aggregation module to address the misalignment issues in scale, region, and task levels. Recently, transformer-based detectors like DETR have made significant progress. Cao \etal \cite{Cao_PSTR_CVPR_2022} introduced a novel transformer-based person search framework. They treated the person search as a set prediction problem and introduced a part attention block to capture the relationship of different parts. 

In this paper, we propose a novel diffusion-based person search framework. Innovatively, we formulate the person search task as a dual denoising process, effectively addressing the limitations of previous approaches by eliminating the prior pedestrian candidates.

\subsection{Diffusion Model}
The class of diffusion-based (or score-based) models has recently become one of the hottest topics in computer vision. Deep diffusion-based generative models \cite{ho2020denoising, song2021scorebased} have demonstrated their remarkable performance in terms of the quality and diversity of the generated samples. So far, diffusion models have been applied to a wide variety of generative tasks, including Image Generation \cite{dhariwal2021diffusion, rombach2022high,jia2024ssmg}, Image Editing \cite{nichol2021glide}, Image Super-Resolution \cite{saharia2022image} \etc. In addition to the generative fields, recent methods also explore the potential of diffusion models for discriminative tasks. Some pioneer works tried to adopt the diffusion model for image segmentation \cite{amit2021segdiff,chen2022generalist,Brempong_2022_CVPR}. Moreover, Han \etal \cite{han2022card} approached supervised learning from a conditional generation perspective and first applied the diffusion model to classification and regression. Chen \etal \cite{chen2022diffusiondet} further adopted the diffusion model for object detection by formulating this task as a generative denoising process. What's more, diffusion models have also shown promising results in detecting anomalies in medical images \cite{pinaya2022fast, wolleb2022diffusion, 9857019}. These excellent works confirm the wide-ranging applicability of diffusion models, indicating that there may be even more untapped applications waiting to be explored.

In general, diffusion models for discriminative tasks follow Noise-to-Label paradigm that supervises learning labels from a conditional generation perspective and denoises the random noise to the ground-truth. This paper extends this paradigm to a novel pipeline (DualNoises-to-DualLabels) by applying the diffusion model to the person search task.


\section{Preliminaries of The Diffusion Model}
\label{sec:preliminaries}
In this section, we introduce the formulation of the diffusion model in detail. Diffusion models are a class of probabilistic generative models, which first gradually add noise to an image until it becomes completely degraded into noise and then learn to reverse the noisy data structure. Thus, the training procedure is divided into two processes: a \textbf{\textit{forward noising process}} and a \textbf{\textit{reverse denoising process}}.

The forward process that gradually adds Gaussian noise to data $x_0 \sim q(x_0)$ over the course of $T$ timesteps
\begin{align}
q(x_{1:T}|x_0) = {\prod^{T}_{t=1} q(x_t|x_{t-1}}),
\end{align}
\begin{align}
q(x_t|x_{t-1}) = \mathcal{N}(x_t;\sqrt{1- \beta_t}x_{t-1}, \beta_tI),
\end{align}
where hyper-parameters $\beta_t$ are set so that $x_T$ is approximately distributed according to a standard normal distribution, $p(x_T)=\mathcal{N}(0,I)$. 
By leveraging the above properties, the reverse process aims to generate new samples from $p_{\theta}\left(x_{0}\right)=\int p_{\theta}\left(x_{0: T}\right) d x_{1: T}$, where 
\begin{align}
p_{\theta}\left(\mathrm{x}_{0: T}\right)=p\left(x_{T}\right) \prod^{T}_{t=1} p_{\theta}\left(\mathrm{x}_{t-1} \mid \mathrm{x}_{t}\right), 
\end{align}

\begin{align}
 p_{\theta}\left(\mathrm{x}_{t-1}\mid \mathrm{x}_{t}\right)=\mathcal{N}\left(\mathrm{x}_{t-1} ; \mu_{\theta}\left(\mathrm{x}_{t}, t\right), \Sigma_{\theta}\left(\mathrm{x}_{t}, t\right)\right),
\end{align}
where $\mu_\theta$ and $\Sigma_\theta$ are the mean and the diagonal covariance matrix to be predicted. Then, this reverse process is trained to match the joint distribution of the forward process by optimizing the variational bound:
\begin{equation}
\begin{aligned}
 L_{\theta}\left(\mathrm{x}_{0}\right)= & \mathbb{E}_{q}\left[\sum_{t>1} D_{\mathrm{KL}}\left(q\left(\mathrm{x}_{t-1} \mid \mathrm{x}_{t}, \mathrm{x}_{0}\right) \| p_{\theta}\left(\mathrm{x}_{t-1} \mid \mathrm{x}_{t}\right)\right) \right. \\
 & \left. +L_{T}\left(\mathrm{x}_{0}\right) -\log p_{\theta}\left(\mathrm{x}_{0} \mid \mathrm{x}_{1}\right) \right],
\end{aligned}
\label{equ:kl}
\end{equation}
where $L_T(x_0) = D_{\mathrm{KL}}(q\left(\mathrm{x}_T, \mathrm{x}_{0}\right) \| p(\mathrm{x}_T))$. The forward process posteriors $q(x_{t-1}|x_t, x_0)$ and marginals $q(x_t|x_0)$ are Gaussian, and the KL divergences in the bound can be calculated in closed form. Thus it is possible to train the diffusion model by taking stochastic gradient steps on random terms of Eq. \eqref{equ:kl}.

\noindent \textbf{Conditional Diffusion Model:} recent researches \cite{saharia2022image,pandey2022diffusevae,ho2022cascaded} proposed the conditional diffusion model as an extension of the above unconditional diffusion models \cite{ho2020denoising,song2021scorebased,sohl2015deep}, which aims to bring the prior knowledge to the reverse denoising process. The conditional diffusion model involves a conditional signal $c$ that is associated with the data $x_0$, such as a label in the context of class-conditional generation, or a low-resolution image in the context of super-resolution \cite{saharia2022image,pandey2022diffusevae,ho2022cascaded}. 
In conditional diffusion models, the goal of the reverse process is to learn a conditional model:
\begin{equation}
p_{\theta }(x_{t-1}|x_t,c) = \mathcal{N}(x_{t-1};\mu_\theta (x_t,c,t),\Sigma_\theta (x_t,c,t)).
\end{equation}

\section{Methodology}
\label{sec:method}

In this section, we begin by reformulating the person search task under the diffusion process. Next, we provide details of our model architecture to explain how to achieve the new formulation. After that, we elaborate the process of training and inference. Finally, we provide some discussions on the comparison between our method and previous methods.

\subsection{Diffusion Process for Person Search}
The person search task tends to train the model to learn pairs $(S, \hat{b}, \hat{e})$, where $S$ indexes scene images and $\hat{b},\hat{e}$ are predicted locations and discriminative embeddings of all possible persons, respectively. It is important to note that the position of each individual in $\hat{b}$ and their corresponding embedding in $\hat{e}$ are directly matched, establishing a one-to-one correspondence. For the sake of simplifying notations, this correspondence relationship is not delineated within this paper.

Based on the aforementioned definitions, we reformulate the task as the diffusion processes, including a dual forward noising process and a dual reverse denoising process. 
\subsubsection{Forward noising process} Given ground truths of bounding boxes $b_0$ and ReID embedding $e_0$, our dual forward noising process $q$ gradually adds Gaussian noise to construct noisy labels. Especially, the forward process is independent of the given condition $c$ as \cite{saharia2022image,pandey2022diffusevae,ho2022cascaded}. The process is specified as follows:
\begin{equation}
\begin{aligned}
q(b_t|b_0,c) = q(b_t|b_0) = \mathcal{N}(b_t;\sqrt{\bar{\alpha}_t}b_{0}, (1-\bar{\alpha}_t)I),\\
q(e_t|e_0,c) = q(e_t|e_0) =  \mathcal{N}(e_t;\sqrt{\bar{\alpha}_t}e_{0}, (1-\bar{\alpha}_t)I),
\end{aligned}
\end{equation}
where $t$ $\sim$ Uniform($\{1,...,T\}$) is the time step sampled from the number of diffusion steps $T$. $\bar{\alpha}_t = {\textstyle \prod_{i=1}^{t}} \alpha_i$ is controlled by the noise variance schedule \cite{nichol2021improved}. $b_t$ and $e_t$ refer to the noisy boxes and embeddings respectively. 

\subsubsection{Reverse denoising process}  Given noisy boxes $b_t$ and noisy ReID embeddings $e_t$, the goal of the reverse denoising process is to learn a conditional model:
\begin{align}
p_{\phi }&(\{b_{t-1},e_{t-1}\}|\{b_t,e_t\},c)  \\ 
&= \mathcal{N}(\{b_{t-1},e_{t-1}\};\mu_\phi (\{b_t,e_t\},c,t),\Sigma_\phi (\{b_t,e_t\},c,t))\notag,
\end{align}
where $\mu_\phi$ and $\Sigma_\phi$ are the mean and diagonal covariance matrix to be predicted \cite{ho2020denoising,songdenoising}. Specifically, $\Sigma_\phi$ is predefined without learning and a neural network $f_{\theta_2}((b_t,e_t),c,t)$ is employed to indirectly approximate $\mu_\phi$ by predicting $\hat{b}_0, \hat{e}_0$ collaboratively. This step-by-step denoising framework enables PSDiff to refine $\hat{b}_0, \hat{e}_0$ in an iterative and collaborative way. In the following, we provide details of our model architecture to explain how to achieve the above procedure.

\subsection{Model Architecture}

\begin{figure*}[t]
    \centering
    \includegraphics[width=0.95\textwidth]{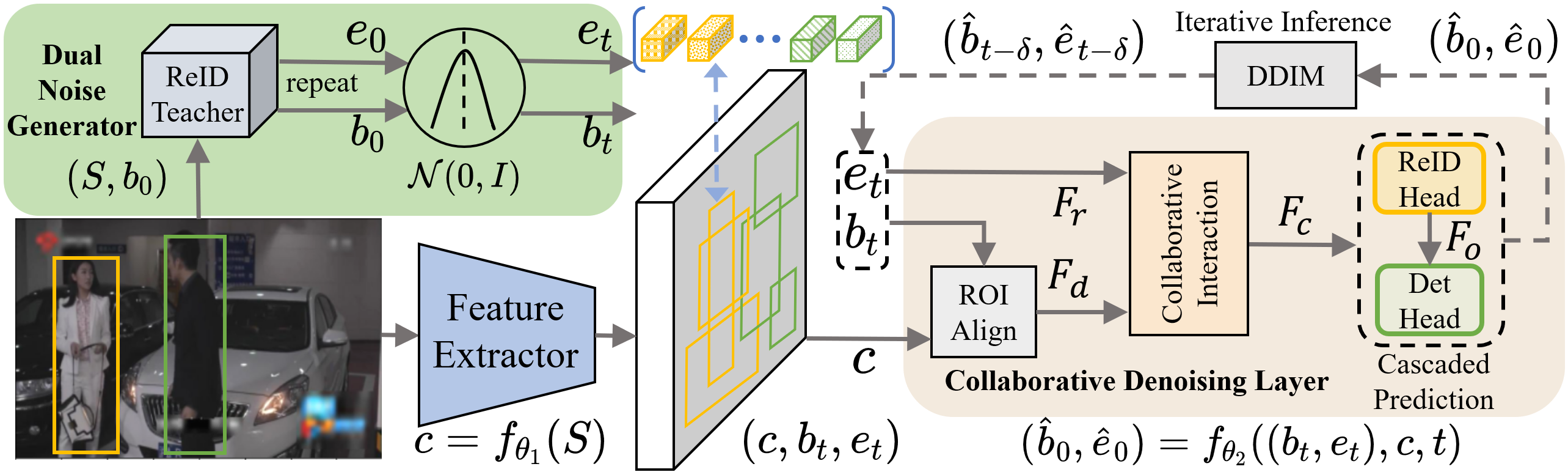}
    \caption{The overall architecture of the proposed PSDiff. During the training stage, $b_t,e_t$ are noisy boxes and embeddings calculated by Eq. \eqref{equ:noisybe}. 
    During the inference stage, $b_t,e_t$ are randomly generated boxes and ReID embeddings from Gaussian noises, which are refined gradually by the iterative inference with fast sampling methods, \eg, DDIM.}
    \label{fig:overview}
\end{figure*}

The overview of the proposed framework is presented in Fig. \ref{fig:overview}. The model consists of three main components: \textit{(1) feature extractor}, \textit{(2) dual noise generator} and \textit{(3) collaborative denoising layer}. The feature extractor first generates conditional features $c$ from the input image. Then the dual noise generator samples the timestep $t$ and transforms ground truths to noisy boxes $b_t$ and noisy embeddings $e_t$. Finally, the collaborative denoising layer receives dual noises and their condition $c,t$ to predict $\hat{b}_0$ and $\hat{e}_0$. We will elaborate on each component in the following subsections.
\subsubsection{Feature Extractor} The feature extractor aims to extract high-level and multi-scale conditional features $c$ from raw scene images $S$. This process is specified as follows:
\begin{align}
\label{equ:backbone}
c = f_{\theta_1}(S).
\end{align}
Specifically, we implement $f_{\theta_1}$ with Feature Pyramid Network (FPN) \cite{lin2017feature} based on both CNN-based ResNet \cite{he2016deep} and Transformer-based Swin \cite{liu2021swin} architecture as \cite{chen2022diffusiondet}.  

\subsubsection{Dual Noise Generator} The dual noise generator is designed to corrupt boxes and embeddings from ground truths to randomly distributed noises. Initially, the ground truth boxes and their associated embeddings are uniformly duplicated until the cumulative total equals a predefined number, \textit{$N_{train}$}, for every image. Then, we sample timesteps $t$ $\sim$ Uniform($\{1,...,T\}$) and corrupt \textit{$N_{train}$} ground truths to the noisy boxes and embeddings by adding Gaussian noises as follows:
\begin{equation}
\begin{aligned}
\label{equ:noisybe}
q(b_t|b_0,c) &= s_1 \sqrt{\bar{\alpha}_t}(b_{0}*2 -1) + \sqrt{(1-\bar{\alpha}_t)}\epsilon_1,  \\
q(e_t|e_0,c) &= s_2 \sqrt{\bar{\alpha}_t}Norm(e_{0}) + \sqrt{(1-\bar{\alpha}_t)}\epsilon_2, 
\end{aligned}
\end{equation}
where $\epsilon_1,\epsilon_2 \sim \mathcal{N}(0,I)$. The noise addition process is applied to each value within every box and every embedding, \eg, with 4 values for each box and 256 values for each embedding. Therefore, operations $b_{0}*2 -1$ and $Norm(e_{0})$ aim to rescale values in ground truths into 
$[-1, 1]$. $s_1$ and $s_2$ are the signal-to-noise ratios (SNR) corresponding  to  $b_0$ and $e_0$, respectively. 
The SNR further adjusts the scaling factor, which has a significant effect on the performance \cite{chen2022generalist, chen2022diffusiondet}. Besides, we keep the noise scalar $\alpha_t$ consistent for both $b_0$ and $e_0$.

\textbf{\textit{How to get the ground truth of embeddings $e_0$?}}
Note that there are only person identities in the original annotations. In this work, we first use annotations of bounding box and identity to train a ReID teacher model as \cite{10234225,hu2021hard}. 
Then, we adopt a strategy of feature imitation to take the output of the ReID teacher model as the ground truth of embeddings $e_0$. 
This process is specified as follows:
\begin{equation}
\label{equ:teacher}
\centering
e_0 = Teacher(S, b_0).
\end{equation}
Note that the ReID teacher model is pre-trained and only used in the training phase of PSDiff. It will be removed during inference and thus will not bring an extra computational burden. Specifically, we employ the supervised HHCL \cite{10234225,hu2021hard} as the teacher model in PSDiff, which offers two distinct advantages. First, it employs hard-sample mining to focus on challenging examples, thereby enhancing the learning process. Second, it leverages unlabeled samples as CCR \cite{10234225} to enrich the training set. These benefits collectively contribute to improving the quality of the learned embeddings, bringing them closer to the ground truth distribution of embeddings.
We next discuss how one can learn a neural network to reverse this Gaussian diffusion process.

\begin{figure}[t]
    \centering
    \includegraphics[width=0.49\textwidth]{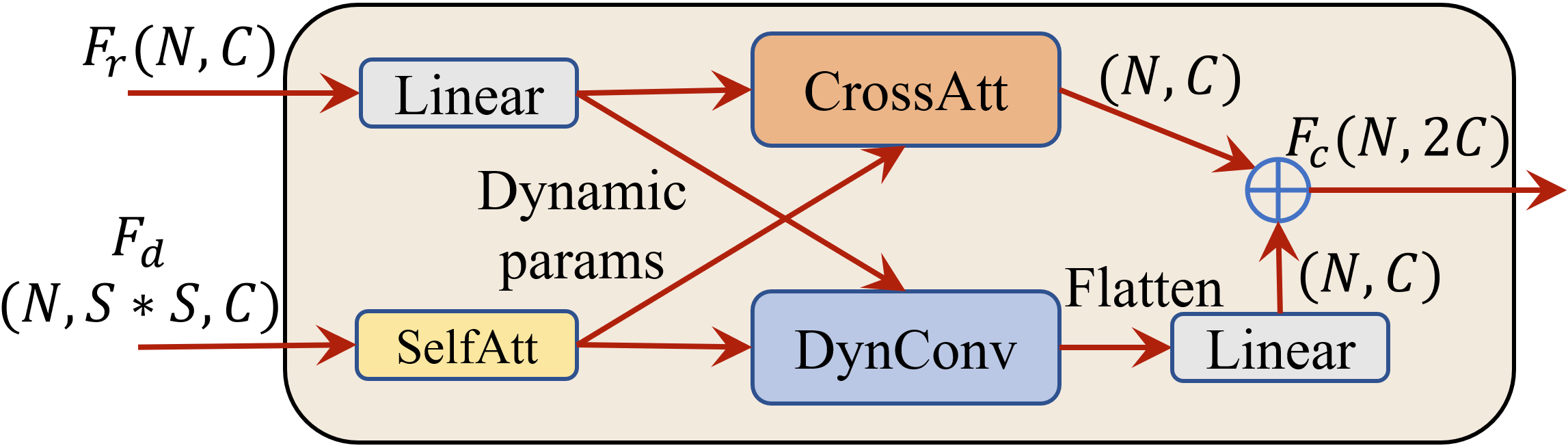}
    \caption{Illustrations of the collaborative interaction.}
    \label{fig:inter}
\end{figure}

\subsubsection{Collaborative Denoising Layer} Given noisy boxes $b_t$ and embeddings $e_t$, we propose the Collaborative Denoising Layer (CDL) to predict $\hat{b}_0, \hat{e}_0$ as:
\begin{equation}
\label{equ:f_cdl}
\centering
(\hat{b}_0, \hat{e}_0) = f_{\theta_2}((b_t, e_t),c,t),
\end{equation}
where $\hat{b}_0, \hat{e}_0$ refer to the predicted boxes and embeddings, respectively.
As shown in Fig. \ref{fig:overview}, CDL first utilizes the RoIAlign operation to extract detection features $F_d(N,S*S,C)$ for $N$ noisy boxes\footnote{Note that we generate \textit{$N_{train}$} noisy boxes for each image, and the number of boxes in one batch is $N=N_{train}*batchsize$.}, where $S,C$ is the size and the channel of features. Then CDL generates the collaborative features $F_c$, by collaboratively interacting $F_d$ and ReID embeddings $F_r(N,C)$ to realize the complementary advantages of the two sub-tasks. 
After the collaborative interaction, $F_c$ is used to predict results. 

\textbf{\textit{(a) Collaborative interaction:}} we present the details of the collaborative interaction in Fig. \ref{fig:inter}. We first apply a linear layer and a  multi-head self-attention module to $F_r$ and $F_d$, separately. The former generates dynamic kernel parameters and the latter aims to model intra-level visual relationships \cite{luo2021dual,sun2021sparse}. Then the new $F_d$ is enhanced by a dynamic convolution, whose kernel parameters are produced by $F_r$. Meanwhile, we apply the cross-attention moudle to $F_r$ and $F_d$, where $F_r$ serves as queries and $F_d$ serve as keys and values. The above collaborative interaction reinforces features of one task by embedding information from the other task. Finally, two kinds of improved features are concatenated as $F_c(N,2C)$ to predict results.

\renewcommand{\algorithmicrequire}{\textbf{Input}}
\renewcommand{\algorithmicensure}{\textbf{return}}
\begin{algorithm}[t]
	\caption{Training process of PSDiff} 
	\label{alg:CGC} 
	\begin{algorithmic}[0] 
		\REQUIRE total diffusion steps $T$, dataset $D$, pretrained ReID teacher network
            \REPEAT
          \STATE 1. Sample data $(S,b_0)$ $\sim$ $D$;
	   \STATE 2. Sample $t$ $\sim$ Uniform($\{1,...,T\}$), $\epsilon_1, \epsilon_2 \sim \mathcal{N}(0,I)$;
         \STATE 3. Compute condition $x$ using $f_{\theta_1}$ by Eq. \eqref{equ:backbone};
         \STATE 4. Compute $e_0$ using teacher network by Eq. \eqref{equ:teacher};
          \STATE 5. Repeat $b_0, e_0$ and compute $b_t, e_t$ by Eq. \eqref{equ:noisybe};
          \STATE 6. Predict $\hat{b}_0, \hat{e}_0$ using $f_{\theta_2}$ by Eq. \eqref{equ:f_cdl};
          \STATE 7. Take gradient step on loss $\mathcal{L}$ by Eq. \eqref{equ:totalloss};  
		\UNTIL convergence
	\end{algorithmic}
\label{alg:train}
\end{algorithm}
\begin{algorithm}[t] 
	\caption{Inference process of PSDiff} 
	\label{alg:test} 
	\begin{algorithmic}[1] 
		\REQUIRE total diffusion steps \(T\), inference interval \(\delta\), image \(S\), \(\hat{b}_T,\hat{e}_T \sim \mathcal{N}(0,I)\)
        \FOR{\(t = T, T-\delta, \cdots, T-k\delta ,\cdots\)}            
            \STATE 1. Compute condition \(x\) using \(f_{\theta_1}\) by Eq. \eqref{equ:backbone};
            \STATE 2. Compute \(\hat{b}_0, \hat{e}_0\) using \(f_{\theta_2}((\hat{b}_t, \hat{e}_t),x,t)\) by Eq. \eqref{equ:f_cdl};
            \IF{\(t > \delta\)}
                \STATE 3. Compute \(\hat{b}_{t-\delta}, \hat{e}_{t-\delta}\) from newly predicted \(\hat{b}_0, \hat{e}_0\) using the fast sampling method;
            \ELSE
                \RETURN \(\hat{b}_0, \hat{e}_0\)
            \ENDIF
        \ENDFOR
	\end{algorithmic}
\end{algorithm}

\textbf{\textit{(b) Cascaded prediction:}} previous detectors always adopt the iterative structure \cite{cai2018cascade,sun2021sparse} to improve the performance, where the newly generated object boxes and object features will serve as the next stage's proposal boxes and proposal features. Object features $F_o$, refer to the high dimensional intermediate features used to predict boxes. 
Sparse R-CNN \cite{sun2021sparse} further confirmed that iteratively optimizing $F_o$ matters 
while only updating the boxes does not make a big difference. However, in our framework, only predicted boxes and ReID embeddings are passed to the next iterative process. Therefore, we design a cascaded architecture instead of a common parallel structure \cite{xiaoli2017joint,li2021sequential} to further refine $F_o$. As shown in Fig. \ref{fig:overview}, the $F_c$ are first used to predict embeddings $\hat{e}_0$. Then we take $F_o = \hat{e}_0$ and feed $F_o$ to the detection head for predicting $\hat{b}_0$. This design makes sense, due to the fact that the dimension of $e_0$ is much larger than $b_0$ in general, 4 for $b_0$ $\ll$ 256 for $e_0$. With the iterative refinement of ReID embeddings, this strategy ensures that both $F_r$ and $F_o$ are rectified, leading to more accurate detection results.

\subsection{Training and Inference}
\subsubsection{Training process}
We adopt an end-to-end training mechanism. The model takes $N_{train}$ noisy boxes and embeddings as input and then predicts $N_{train}$ box coordinates and ReID embeddings for each image.
The pseudo-code of the detailed training process is shown in Algorithm \ref{alg:train}. In this work, we extend the set prediction loss \cite{carion2020end,stewart2016end} to the person search task for training our diffusion-based model. The extended loss produces an optimal bipartite matching between predictions and ground truths as previous methods \cite{carion2020end,stewart2016end}. 
Differently, the matching cost is defined as
\begin{equation}
\label{equ:totalloss}
\centering
\mathcal{L} = \underbrace{\lambda_{cls} \mathcal{L}_{cls} + \lambda_{L1} \mathcal{L}_{L1} + \lambda_{giou}  \mathcal{L}_{giou}}_{denoise: b_t} + \underbrace{\lambda_{reid}  \mathcal{L}_{reid}}_{denoise:  e_t},
\end{equation}
where $\lambda_{cls}, \lambda_{giou}, \lambda_{L1}$ and $\lambda_{reid}$ are coefficients of each component. $\mathcal{L}_{cls}, \mathcal{L}_{L1}$, and $\mathcal{L}_{giou}$ denote the focal loss \cite{lin2017focal}, $\ell_1$ loss, and generalized IoU loss \cite{rezatofighi2019generalized}, respectively. We adopt the generalized IoU (GIoU) loss to address the limitations of standard IoU in handling non-overlapping bounding boxes, which improves the optimization process for more accurate bounding box regression. For a more detailed understanding of GIoU, we encourage readers to refer to the original work \cite{rezatofighi2019generalized}. We calculate these losses as the previous methods \cite{carion2020end,sun2021sparse} of object detection. In addition, losses for denoising $e_t$ are defined as
\begin{align}
\label{equ:loss_reid}
\mathcal{L}_{reid} =  \sum_{i=1}^M \|e_0^i-\hat{e}_0^{\sigma(i)}\|_2,
\end{align}
where $i$ is the index of ground truths and $\sigma(i)$ is the index of bipartite matching results corresponding to $i$. $M$ is the number of ground truths for each image. We employ the $\ell_2$ loss to supervise predicted ReID embeddings $\hat{e}_0$ with $e_0$.

\subsubsection{Inference via Iterative Refinement} The inference procedure of PSDiff is a denoising process, starting from $N_{test}$ Gaussian noises $\hat{b}_T,\hat{e}_T \sim \mathcal{N}(0,I)$. The model refines predictions in an iterative way under the sampling strategy, $(\hat{b}_T,\hat{e}_T) \rightarrow (\hat{b}_{T-\delta},\hat{e}_{T-\delta}) \rightarrow ... \rightarrow (\hat{b}_{0},\hat{e}_{0})$, where $\delta$ is the inference interval. In each step, the refined boxes and embeddings from the last step are sent into the CDL to denoise gradually. However, traditional step-by-step inference usually needs thousands of sequential steps of CDL evaluations, thereby giving rise to a much slower inference speed than previous person search methods. Therefore, we employ some fast sampling methods, \eg, DDIM \cite{songdenoising}, DPM-solver \cite{ludpm} and DPM-solver++ \cite{lu2022dpm} to accelerate the inference. The pseudo-code of the inference process is shown in Algorithm \ref{alg:test}.

\subsection{Discussion}
In this section, we provide some discussions, primarily focusing on the distinctions and comparisons between our method and previous approaches. Specifically, we compare our diffusion-based framework with methods that are built upon Faster-RCNN or DETR architectures. Additionally, we provide a detailed exposition of the distinctions between our methodology and DiffusionDet \cite{chen2022diffusiondet}. What's more, we explain how our approach stands in relation to methods that employ knowledge distillation techniques. Finally,  We discuss the impact and prospect of \textit{pretrained backbones on large-scale multi-modal person datasets} for person search.

\subsubsection{\textbf{Comparison to Non-diffusion based methods}}
previous person search approaches have traditionally relied on established detectors that involve predefined object candidates, \eg, empirical object priors in Faster-RCNN \cite{rennips15fasterrcnn} and learnable object queries in DETR \cite{carion2020end}. These hand-designed or pre-learned object candidates inadvertently prioritize object detection over ReID, leading to sub-optimal performance in the ReID task and hindering effective collaboration between the two tasks. In contrast, our diffusion-based method eliminates such prior pedestrian candidates.  It's worth noting that while DMRNet\cite{han2021decoupled} and RDLR\cite{han2019re} have mentioned the conflict and potential collaboration between the two sub-tasks, they differ significantly from our approach. DMRNet aims to avoid the conflict of subtasks, but they do not further explore collaboration between the tasks. RDLR suggests that ReID can guide better bounding box predictions, but their primary focus is on mitigating detector bias in two-stage methods, lacking an end-to-end unified model. In contrast, our method enables collaborative optimization for the entire end-to-end person search model, representing a significant improvement over these earlier approaches.  We note a recent work \cite{chen2023transferring} on Scene Adaptation shares the similar idea of leveraging multi-task collaboration to improve overall performance. In their case, segmentation and depth estimation are treated as complementary tasks, where depth information helps guide segmentation and vice versa. Similarly, in our approach, we enable collaborative optimization between pedestrian detection and re-identification (ReID). 
This highlights that collaborative learning between tasks is crucial and can be applied across various domains.

\subsubsection{\textbf{Comparison to DiffusionDet}} DiffusionDet \cite{chen2022diffusiondet}, a notable object detection model, skillfully mitigates the reliance on prior candidates through its innovative diffusion model. We respectfully acknowledge the foundational benefits derived from DiffusionDet, yet it is crucial to underscore that the domain of person search introduces a considerably more intricate challenge than traditional object detection. This complexity stems from the imperative to simultaneously and effectively execute both detection and Re-Identification (ReID) tasks. The essence of our innovation lies in the formulation of a diffusion architecture, meticulously engineered to address these tasks concurrently and cultivate a synergistic collaboration between them. In response to these sophisticated challenges, we have innovated the Dual Noise Generator and Collaborative Interaction mechanisms. Our methodology advances beyond DiffusionDet's initial framework and ingeniously augments DiffusionDet's principal strengths to adeptly cater to the complex demands of person search tasks, showcasing our approach's novelty and superiority.

\subsubsection{\textbf{Pretrained backbones on large-scale multi-modal person datasets }}
Recent advancements \cite{yang2023towards,he2024instruct} in large person datasets aim to introduce foundational models that utilize multimodal information, such as pairing textual attributes or descriptions with pedestrian data. These datasets are particularly valuable for tasks like multimodal re-identification, where models benefit from learning cross-modal features. In this paper, person search is a purely visual retrieval task, and multimodal pretraining does not yet provide clear advantages in this context. Nevertheless, using pre-trained backbones from these large multimodal datasets remains a highly promising direction for future work. 

\section{Experiments}
In this section, we empirically evaluate the performance of
PSDiff on two benchmark datasets. We also provide detailed ablation studies to illustrate how each component of our method works.  Furthermore, we conduct case studies to better illustrate the effectiveness of our method.
\label{sec:exp}
\subsection{Experimental Setup}
\subsubsection{Dataset}
We empirically evaluate the effectiveness of the proposed framework over two standard benchmarks.
\begin{itemize}
\item {\bf CUHK-SYSU \cite{xiaoli2017joint}} is a large-scale dataset that is collected from street snaps or movies, including 18,184 images and 96,143 annotated pedestrian bounding boxes (23,430 boxes are ID labeled from 8,432 identities). These images cover significant variations of viewpoints, lighting and background conditions.
The training set consists of 11,206 images from 5,532 identities, while the testing set contains 2,900 query images and 6,978 gallery images. We report the results with the default gallery size of 100 if not specified.
\item {\bf PRW \cite{Zheng_2017_CVPR}} is extracted from six surveillance cameras on a university campus, including 11,816 video frames and 43,110 bounding boxes (34,304 boxes are ID labeled from 932 identities). PRW is more challenging because each identity has more instances (36.8 \vs~2.8 in CUHK-SYSU), making retrieving difficult.  The training set consists of 5,704 frames
and 482 identities, while the testing set includes 6,112 gallery
images and 2,057 query images from 450 different identities. 
\end{itemize}

\subsubsection{Evaluation Protocol}
We report the performance of the person search in terms of the Cumulative Matching Characteristic (CMC) and the mean Average Precision (mAP). 
CMC is the most popular evaluation metric for re-identification, where matching is considered correct only if the IoU between the ground truth
bounding box and the matching box is larger than 0.5. The mAP reflects the accuracy of localizing the query in the gallery, where the AP score of searching a query from gallery images is calculated first and then AP scores are averaged across all the queries to calculate the mAP. We also report the detection performance under the mAP50 metric.

\subsubsection{Implementation Details} The feature extractor in our network adopts the ResNet-50 \cite{he2016deep} and Swin-Base \cite{liu2021swin} pretrained on the ImageNet dataset, together with the FPN \cite{lin2017feature}.  The dual noise generator adopts the monotonically decreasing cosine schedule for $\alpha_t$ in different timesteps as in \cite{nichol2021improved}. The denoising layer (CDL) is initialized with Xavier init \cite{glorot2010understanding}.  For the ReID teacher model, We strictly adhere to the settings outlined in HHCL \cite{hu2021hard} and CCR \cite{10234225}. We train the PSDiff using AdamW \cite{loshchilov2017decoupled} optimizer with the initial learning rate as $2.5\times10^{{-5}}$ and the weight decay as $10^{{-4}}$. The default training schedule is $90K$ iterations, with the learning rate divided by 10 at $70K$ and $84K$ iterations. The input images are resized to $1333\times800$ by default. If not specified, $s_1$, $ s_2$ and $\lambda_{reid}$ are set to 2.0, 3.0 and 5.0, respectively. Both $N_{train}$ and $N_{test}$ are set to 300. We employ DDIM as the fast sampling method and inference steps are 8 by default. Following \cite{carion2020end,sun2021sparse,chen2022diffusiondet}, $\lambda_{cls}=2,\lambda_{L1}=5,\lambda_{giou}=2$.
The network is trained with a mini-batch size 32 on 4 NVIDIA V100 and all experiments are implemented on the PyTorch framework.

\setlength{\belowcaptionskip}{-0.0cm}
\begin{table}[t]
  \centering
  \tabcolsep = 7 pt
  \caption{Comparison of mAP and top-1 performance with the state-of-the-art methods on CUHK-SYSU and PRW.}\label{tab:SOTA}
  \begin{tabular}{l|l|cc|cc}
    \toprule
    \multicolumn{2}{c|}{\multirow{2}{*}{Methods}} & \multicolumn{2}{c|}{CUHK} & \multicolumn{2}{c}{PRW} \\
    \multicolumn{2}{c|}{}  & mAP & top-1 & mAP & top-1 \\
    \midrule
    \multirow{6}{*}{\rotatebox{90}{two-stage}} & DPM(CVPR'17)\cite{Zheng_2017_CVPR}  & - & - & 20.5 & 48.3 \\
    & RCAA(ECCV'18)\cite{chang2018rcaa}    & 79.3 & 81.3 &   -  &  -   \\
                            & RDLR(ICCV'19)\cite{han2019re}       & 93.0 & 94.2 & 42.9 & 70.2 \\

                            & MGTS(TIP'20)\cite{9003518} & 83.0 & 83.7 & 32.6 & 72.1 \\
                            & IGPN(CVPR'20) \cite{dong2020instance} & 90.3 & 91.4 & 47.2 & 87.0 \\
                            & TCTS(CVPR'20)\cite{wang2020tcts}    & \textbf{93.9} & \textbf{95.1} & 46.8 & 87.5 \\
                            & OR(TIP'21)\cite{9265450}    & 92.3 & 93.8 & 52.3 & 71.5 \\
                            & CCR(TIP'23)\cite{10234225}    & 93.1 & 93.7 & \textbf{54.5} & \textbf{89.5} \\
                            
    \midrule 
    \midrule                     
    \multirow{26}{*}{\rotatebox{90}{end-to-end}} &  OIM(CVPR'17)\cite{xiaoli2017joint}   & 75.5 & 78.7 & 21.3 & 49.4 \\
                            
                            & CTXG(CVPR'19)\cite{Yan_2019_CVPR}   & 84.1 & 86.5 & 33.4 & 73.6 \\
                            & KD(BMVC'19)\cite{munjal2019knowledge}       & 93.0 & 94.2 & 42.9 & 70.2 \\
                            & HOIM(AAAI'20)\cite{chen2020hoim}     & 89.7 & 90.8 & 39.8 & 80.4 \\                      
                            & BINet(CVPR'20)\cite{Dong_2020_CVPR}  & 90.0 & 90.7 & 45.3 & 81.7 \\
                            & NAE+(CVPR'20)\cite{chen2020norm}      & 92.1 & 92.9 & 44.0 & 81.1 \\
                            & EGKD(T CYBER'21) \cite{8759990}
                            &91.1 & 91.9 &34.5 &59.9 \\
                            & HDL(PR'21)\cite{han2021decoupled}      &86.0 & 87.3 &33.6 &69.2 \\
                            & DKD(AAAI'21)\cite{zhang2021diverse}  & 93.1 & 94.2 & 50.5&87.1\\
                              & PGA(CVPR'21)\cite{Kim_2021_CVPR} &  90.2 &  91.8 & 42.5 & 83.5 \\
                            & AGWF(ICCV'21)\cite{Han_2021_ICCV} &93.3&94.2&53.3&87.7\\
                           & SeqNet(AAAI'21)\cite{li2021sequential} & 93.8 & 94.6 & 46.7 & 83.4 \\
                           & Align(CVPR'21)\cite{DBLP:conf/cvpr/YanLQBL00021}  &  93.1 & 93.4 & 45.9 & 81.9 \\
                            & BUFF(TCSVT'21)\cite{9352705}  &  91.6 &  92.2 & 44.9 & 86.3 \\
                           & IIDFC(TCSVT'21)\cite{9438639}  &  93.3 &  92.0 & 43.4 & 83.4 \\
                           & DMRNet(AAAI'21)\cite{han2021decoupled}      &93.2 &94.2 &46.9 &83.3 \\
                           & PSTR(CVPR'22)\cite{Cao_PSTR_CVPR_2022}  &  93.5 & 95.0 & 49.5 & \textbf{87.8} \\
                            & COAT(CVPR'22)\cite{yu2022coat} &  94.2 &  94.7 & 53.3 & 87.4 \\
                            & OIMN(ECCV'22)\cite{lee2022oimnet++}  &  93.1 &  93.9 & 46.8 & 83.9 \\
                            & CANR(TCSVT'22)\cite{9785793}  &  93.9 &  94.5 & 44.8 & 83.9 \\
                            & InvarPS(ICASSP'23) \cite{10095679}  &  93.5 &  94.3 & 52.3 & 85.1 \\
                            & ROICoseg(PR'23)\cite{ZHANG2024110053}  &  92.3 &  93.5 & 50.9 & 86.2 \\

                            &SeqNetXt(WACV'23)\cite{jaffe2023gallery}       & 94.1 & 94.7 & 50.8 & 86.0 \\
                            &KCD (TCSVT'24)\cite{10599306} 
                              & 88.1 & 89.5 & 44.6&84.4\\

                            & SparseRCNN(step-0)\cite{sun2021sparse}       & 90.5 & 92.3 & 49.1 & 80.9 \\
                            
      &\cellcolor{color1}PSDiff (step-1)   & \cellcolor{color1}90.8 & \cellcolor{color1}92.1 & \cellcolor{color1}49.9 & \cellcolor{color1}80.4 \\
      &\cellcolor{color1}PSDiff (step-2)   & \cellcolor{color1}92.5 & \cellcolor{color1}93.8 & \cellcolor{color1}51.1 & \cellcolor{color1}83.2 \\      
      &\cellcolor{color1}PSDiff (step-4)   & \cellcolor{color1}94.7 & \cellcolor{color1}95.1 & \cellcolor{color1}52.7 & \cellcolor{color1}86.3 \\
      &\cellcolor{color1}PSDiff (step-8)   & \cellcolor{color1}\textbf{95.1} & \cellcolor{color1}\textbf{95.3} & \cellcolor{color1}\textbf{53.5} & \cellcolor{color1}87.1 \\
    \cmidrule{2-6}
    & Improved backbone \\
    & PGA-Dilation \cite{Kim_2021_CVPR} &  92.3 &  94.7 & 44.2 & 85.2 \\
    & PSTR-PVTv2 \cite{Cao_PSTR_CVPR_2022} &  95.2 & 96.2 & 56.5 & \textbf{89.7} \\
    & OIMN-Norm \cite{lee2022oimnet++} &  93.1 &  94.1 & 47.7 & 84.8 \\
    & Align-DC \cite{DBLP:conf/cvpr/YanLQBL00021} &  94.0 & 94.5 & 46.1 & 82.1 \\
    & Align*(AAAI'24) \cite{tian2023divide} &  95.4 &  96.0 & 54.5 & 87.6 \\
    & \cellcolor{color1}PSDiff-Swin (step-8)   & \cellcolor{color1}\textbf{95.7} & \cellcolor{color1}\textbf{96.3} & \cellcolor{color1}\textbf{57.1} & \cellcolor{color1}88.1 \\
    \bottomrule
  \end{tabular}
  
\end{table}
\setlength{\belowcaptionskip}{-0.0cm}
\begin{table}[t]
  \centering
  \tabcolsep = 16 pt
  \caption{Complexity and inference speed comparisons.\label{tab:runtime}}
  \begin{tabular}{l|c|c|c}
    \toprule
    Methods & Params & Flops & FPS\\ 
    \midrule
     NAE  \cite{chen2020norm}  & 33M & 575G & 14 \\
     SeqNet  \cite{li2021sequential}  & 48M & 550G & 12\\
     AlignPS  \cite{DBLP:conf/cvpr/YanLQBL00021} & 42M & 380G & 16\\
     RDLR  \cite{han2019re}  & - & - & 15\\
     PSTR  \cite{Cao_PSTR_CVPR_2022}  & 43M & 356G & 18\\
     COAT  \cite{yu2022coat}  & 37M & 473G & 11\\
      \rowcolor{color1} PSDiff (step-1)  & 31M & 301G & 31\\
      \rowcolor{color1} PSDiff (step-2)  & 31M & 309G & 27\\
      \rowcolor{color1} PSDiff (step-4)  & 31M & 333G & 21\\
      \rowcolor{color1} PSDiff (step-8)  & 31M & 390G & 14\\
     
     \midrule    
  \end{tabular}%
\end{table}


\subsection{Comparison to the State-of-the-art}
In this section, we compare the proposed method with current state-of-the-art (SOTA) methods, including two-stage and end-to-end, on the two benchmark datasets. For the two-stage approach, we choose
8 baselines that exhibit excellent performance, including DPM \cite{Zheng_2017_CVPR}, RCAA \cite{chang2018rcaa}, RDLR \cite{han2019re}, MGTS \cite{9003518},  IGPN \cite{dong2020instance}, TCTS \cite{wang2020tcts}, OR \cite{9265450} and CCR \cite{10234225}.
We also compare our methods with 13 end-to-end baselines,
such as NAE \cite{chen2020norm}, AlignPS \cite{DBLP:conf/cvpr/YanLQBL00021}, COAT \cite{yu2022coat}, \etc. 

\begin{figure}[t]
  \centering
     \centering
     \includegraphics[width=0.38\textwidth]{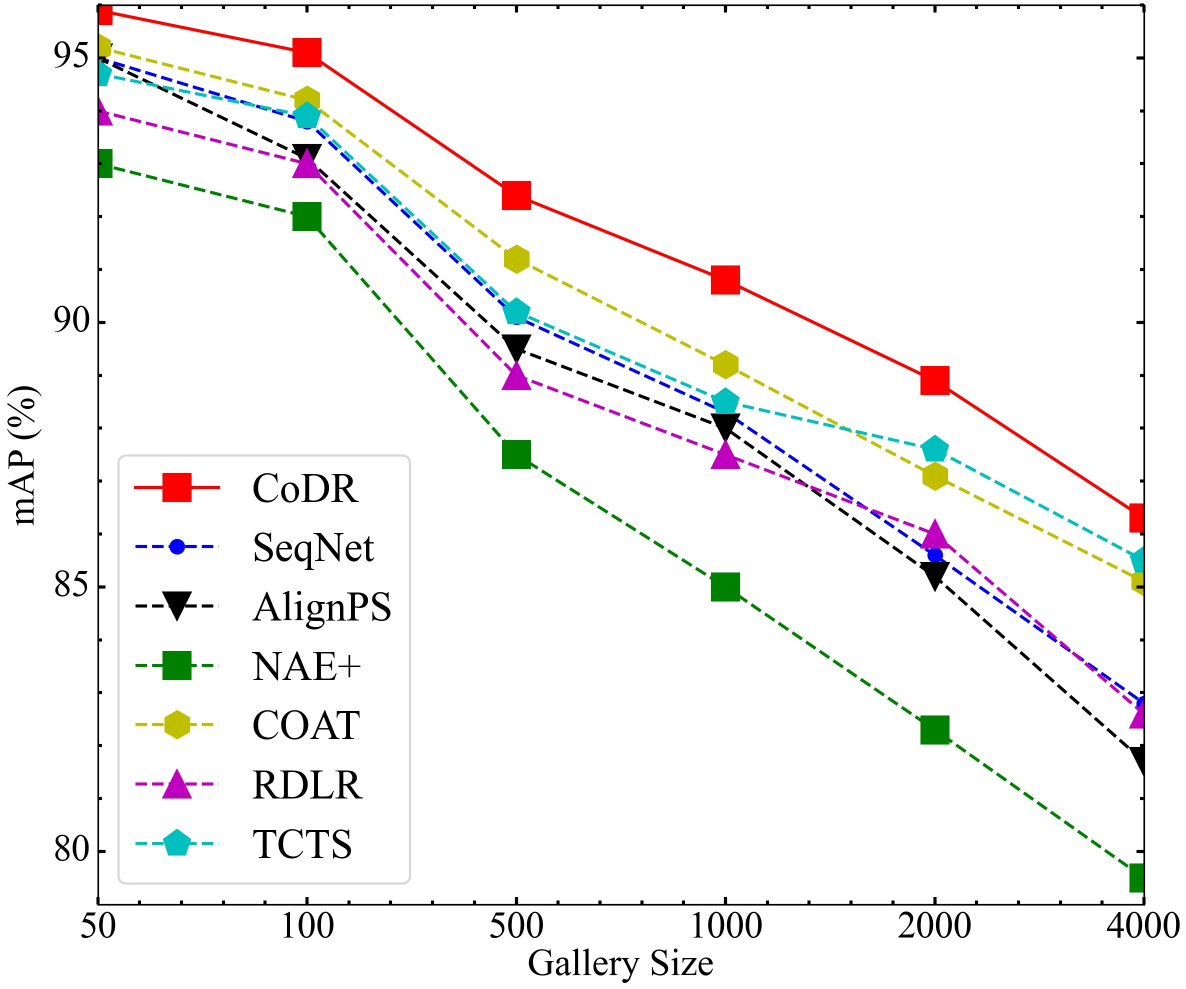}
   \caption{Comparison to different methods on CUHK-SYSU under varying gallery sizes.}
   \label{fig:gallery}
\end{figure}

\setlength{\belowcaptionskip}{-0.0cm}
\begin{table}[t]
  \centering
  \tabcolsep = 13 pt
  \caption{Comparison of transferability with other methods.}\label{tab:Transferability}
  \begin{tabular}{l|cc|cc}
    \toprule
    \multirow{2}{*}{Methods} & \multicolumn{2}{c|}{CUHK$\rightarrow$PRW} & \multicolumn{2}{c}{PRW$\rightarrow$CUHK} \\
     & mAP & top-1 & mAP & top-1 \\
    \midrule
     SeqNet \cite{li2021sequential} & 31.4 & 79.1 & 55.1 & 57.3 \\
     AlignPS \cite{DBLP:conf/cvpr/YanLQBL00021}  & 27.9 & 77.4 & 43.6 & 47.2 \\
     COAT \cite{yu2022coat}       & 29.0 & 77.8 & 56.9 & 59.4 \\
     DAPS \cite{li2022domain} & 30.3 & 77.7 & 52.5 & 54.8 \\     
    \rowcolor{color1} PSDiff (step-8)   & \textbf{35.5} & \textbf{82.1} & \textbf{58.2} & \textbf{63.7} \\
    \bottomrule
  \end{tabular}
\end{table}

\begin{figure*}[t]
  \centering
   \includegraphics[width=0.98\linewidth]{{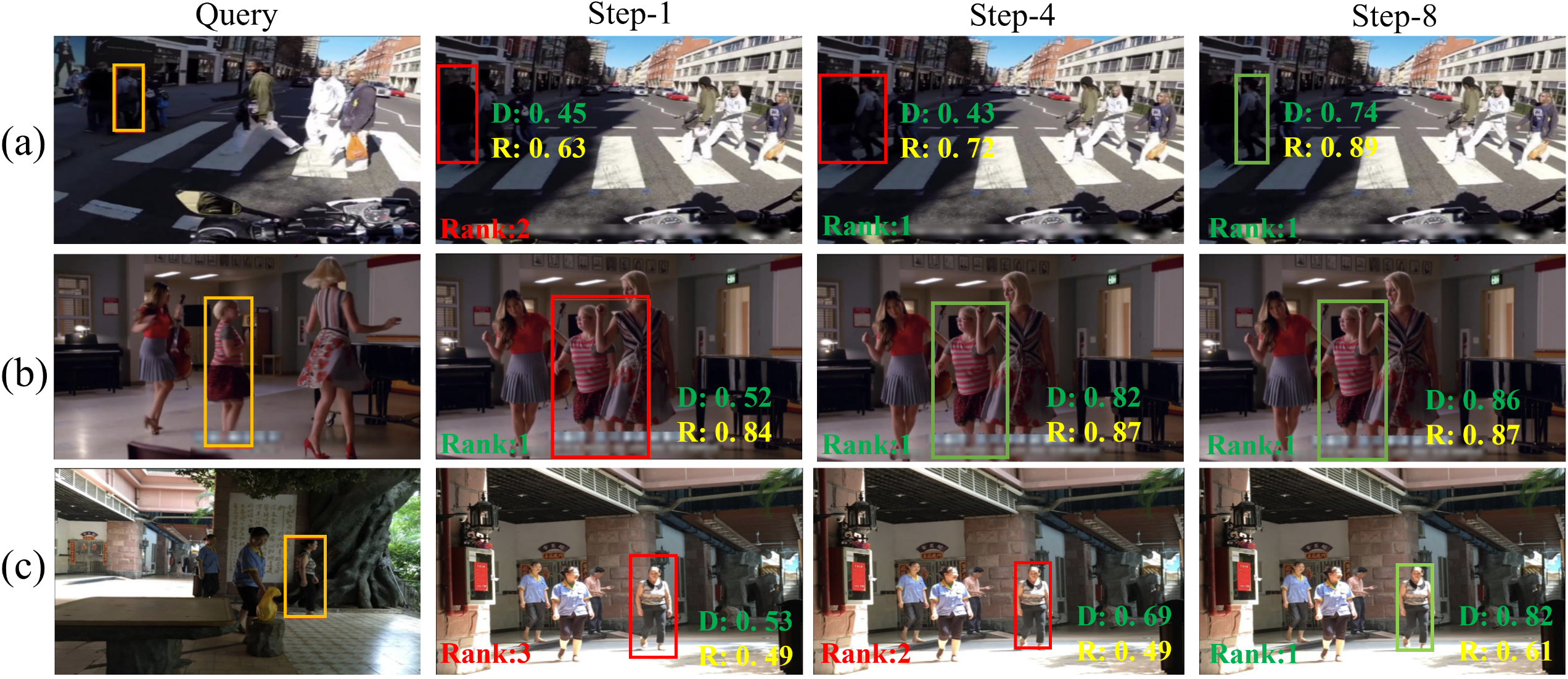}}
   \caption{Qualitative search results on the CUHK-SYSU dataset. The bounding boxes in yellow denote the queries while green and red denote the correct and wrong results.``D'' and ``R'' index confidence scores of detection and ReID matching scores with queries, respectively. ``Rank-n'' refers to the rank of the presented results among all predictions of the gallery.}
   \label{fig:Case}
\end{figure*}

\begin{table*}[t]
\tabcolsep = 7 pt
\centering
\caption{Ablation experiments of the proposed PSDiff. (a) \textbf{Signal-to-Noise Ratios}: $s_1,s_2$ are the SNR corresponding to detection and ReID tasks. (b) \textbf{Collaborative Interaction}: ``SA", ``DC" and ``CA" refer to the self-attention module, dynamic convolution and cross-attention module in collaborative interaction, respectively. (c) \textbf{Cascaded Prediction.}}
\label{tab:ablation}
\subfloat[]{
\begin{tabular}{c|cccc}
    \toprule
    \rule{0pt}{10pt}
    \diagbox[]{$s_1$}{$s_2$}  & 1.0 & 2.0 & 3.0  & 4.0 \\ [1pt]
    \midrule    
    1.0 & 93.9 & 94.0 & 94.3 & 93.3 \\
    2.0 & 94.2 & 94.6 & \cellcolor{color1}{\textbf{95.1}} & 94.4 \\
    3.0 & 94.1 & 94.2 & 94.5 & 94.1 \\
    4.0 & 93.8 & 94.0 & 94.1 & 93.8 \\
    \bottomrule
\end{tabular}    
\label{tab:ablationa}
}
\hspace{0.3cm}
\subfloat[]{
\begin{tabular}{ccc|cc}
    \toprule
    \tabcolsep=7 pt
    SA & DC & CA  & mAP & top-1  \\
    \midrule    
      & & & 91.3 & 91.8  \\
    $\surd$ &  &  & 92.5 & 93.3  \\
    $\surd$ & $\surd$ &  & 93.8 & 94.3  \\
    $\surd$ &  & $\surd$  & 94.3 & 94.7  \\
    \rowcolor{color1} $\surd$  & $\surd$ & $\surd$ & \textbf{95.1} & \textbf{95.3}  \\
    \bottomrule
\end{tabular}
\label{tab:ablationb}
}
\hspace{0.3cm}
\subfloat[]{
\renewcommand{\arraystretch}{1.2}
\tabcolsep=10 pt    
\begin{tabular}{c|cc}
    \toprule
    Methods  & mAP & top-1  \\
    \midrule    
    parallel(step-1) & 90.5 & 91.7  \\
    parallel(step-8) & 93.5 & 94.1  \\
    cascaded(step-1) & 90.8 & 92.1\\
    \rowcolor{color1} cascaded(step-8) & \textbf{95.1} & \textbf{95.3}  \\
    \bottomrule
\end{tabular}    
\label{tab:ablationc}
}
\end{table*}

\begin{table}[t]
  \centering
    \tabcolsep = 6 pt
\renewcommand\arraystretch{1.3}
\caption{Comparative results using different inference strategies, such as DDIM, DPM-solver and DPM-solver++.}
\begin{tabular}{c|cccc}
 			\toprule
 			Methods  & step-1 & step-2 & step-4 &  step-8 \\
 			\midrule
                 DDIM& 90.8 & 92.5 & 94.7 & \textbf{95.1} \\
\rowcolor{color1}DPM-solver & 90.8 & \textbf{92.8} & \textbf{94.9} & \textbf{95.1} \\
                 DPM-solver++ & 90.8 & 92.5 & 94.8 & 95.0 \\
 			\bottomrule
 		\end{tabular}            
            \label{tab:sampling}
\end{table}
\setlength{\belowcaptionskip}{-0.0cm}

\subsubsection{Evaluation On CUHK-SYSU dataset} 
Table \ref{tab:SOTA} shows the performance comparisons on the CUHK-SYSU dataset. The proposed PSDiff achieves 90.8\% mAP without refinement (step-1), which has a performance margin with SOTA methods. However, the performance of PSDiff can be improved significantly with iterative and collaborative refinements.  As shown in the gray block of Table \ref{tab:SOTA}, our PSDiff achieves 95.1\% mAP and 95.3\% top-1 accuracy with 8 iterative steps, outperforming existing SOTA methods. Moreover, some methods employ improved backbones for better performance, such as R50-Dilation in \cite{Kim_2021_CVPR}, deformable convolution (DC) in \cite{DBLP:conf/cvpr/YanLQBL00021}, transformer in \cite{Cao_PSTR_CVPR_2022} and ProtoNorm in \cite{lee2022oimnet++}. For a fair comparison, we also replace ResNet50 with Swin-B \cite{liu2021swin} to boost performance. As shown in the bottom of Table \ref{tab:SOTA}, PSDiff achieves 95.7\% mAP and 96.3\% top-1 accuracy, outperforming strong methods such as Align-DC \cite{DBLP:conf/cvpr/YanLQBL00021} and PSTR-PVTv2 \cite{Cao_PSTR_CVPR_2022}.

To further validate the contribution of the diffusion part in PSDiff, we conducted an experiment where we removed the diffusion component while retaining the original attention modules. The resulting model is based on the Sparse R-CNN structure \cite{sun2021sparse}, with two separate branches for detection and ReID, which we refer to as the ``step-0" method. As shown in the results, while the ``step-0" method achieves similar performance to our diffusion model at step-1, it lacks the ability to perform iterative inference, which highlights the importance of the diffusion mechanism.

To evaluate the performance scalability of methods, we also compare
with other competitive methods under varying gallery sizes of $[50,100,500,1000,2000,4000]$ in Fig. \ref{fig:gallery}. A larger gallery size corresponds to a larger search scope, meaning that more distracting people are involved in matching, which is closer to real-world applications. When the gallery size increases, our method still outperforms all existing methods, indicating that our method can handle more challenging real-world situations.

\subsubsection{Evaluation On the PRW dataset}
We also show comparisons on the PRW dataset in Table \ref{tab:SOTA}. PRW is more challenging because of less training data and a larger gallery size, which causes poor performance of mAP in existing methods. 
Surprisingly, our proposed PSDiff reaches 
the best performance on mAP, 53.5\% with ResNet-50 and 57.1\% with Swin-Base, demonstrating its efficacy under various scenarios. We also note that PSTR \cite{Cao_PSTR_CVPR_2022} obtains the best performance on top-1 accuracy. However, it needs to train three detection decoders and a ReID shared decoder with the multi-level supervision scheme, suffering from the high parameter complexity. In contrast, our PSDiff only trains a single prediction layer and the new paradigm enables us to reuse network parameters.


\subsubsection{Comparison to methods with knowledge distillation}
Previous approaches \cite{munjal2019knowledge,8759990,han2021decoupled,zhang2021diverse,10599306} pointed out that the performance bottleneck in end-to-end person search models arises from the feature map containing redundant contextual information. To mitigate this issue, they employ Knowledge Distillation (KD) to provide superior prior knowledge. In our work, we adopt a similar strategy by leveraging the output of the pre-trained ReID teacher to approximate the ground truth of embeddings. 
Although the form of our loss function shares similarities with KD-based methods, our approach offers a unique advantage: it enables iterative refinement of the predicted embeddings during the inference stage. Experimental results in Table \ref{tab:SOTA}  clearly show that our method outperforms the previous KD-based approaches.

	\subsubsection{Efficiency Comparison} To show the efficiency properties of our method, we present the complexity of parameters (Params) and computation (Flops) in Table \ref{tab:runtime}. We also report the inference speed (FPS) for a fair comparison, computed on the same Tesla V100 GPU. As shown in Table \ref{tab:runtime}, PSDiff requires the fewest parameters since only a single denoising layer is trained. Moreover, unlike traditional fixed models that rely on larger backbones or deeper layers to improve performance, PSDiff takes advantage of the diffusion model’s ability to reuse trained layers. This enables elastic computational overhead, allowing users to adjust the number of inference steps to balance the trade-off between speed and accuracy. These advantages of efficiency make the model more scalable and adaptable to varying computational resources.


\subsubsection{Transferability Comparison} As illustrated in Table \ref{tab:Transferability}, we assess the model's performance across datasets without any additional fine-tuning. The results show that previous methods suffer from a severe decline in performance when applied to new datasets. In contrast, our approach sets a new standard by achieving the best results in cross-dataset evaluations. These findings demonstrate that our diffusion-based architecture achieves not only improved performance within the dataset but also exhibits exceptional transferability across different datasets. The remarkable transferability of DiffPS suggests that it serves as an effective method for person search tasks across a wide range of scenarios, \eg, from sparsely populated areas to densely crowded environments, all without necessitating further fine-tuning.

\begin{figure}[t]
     \hspace{-0.5cm}
     \includegraphics[width=0.515\textwidth]{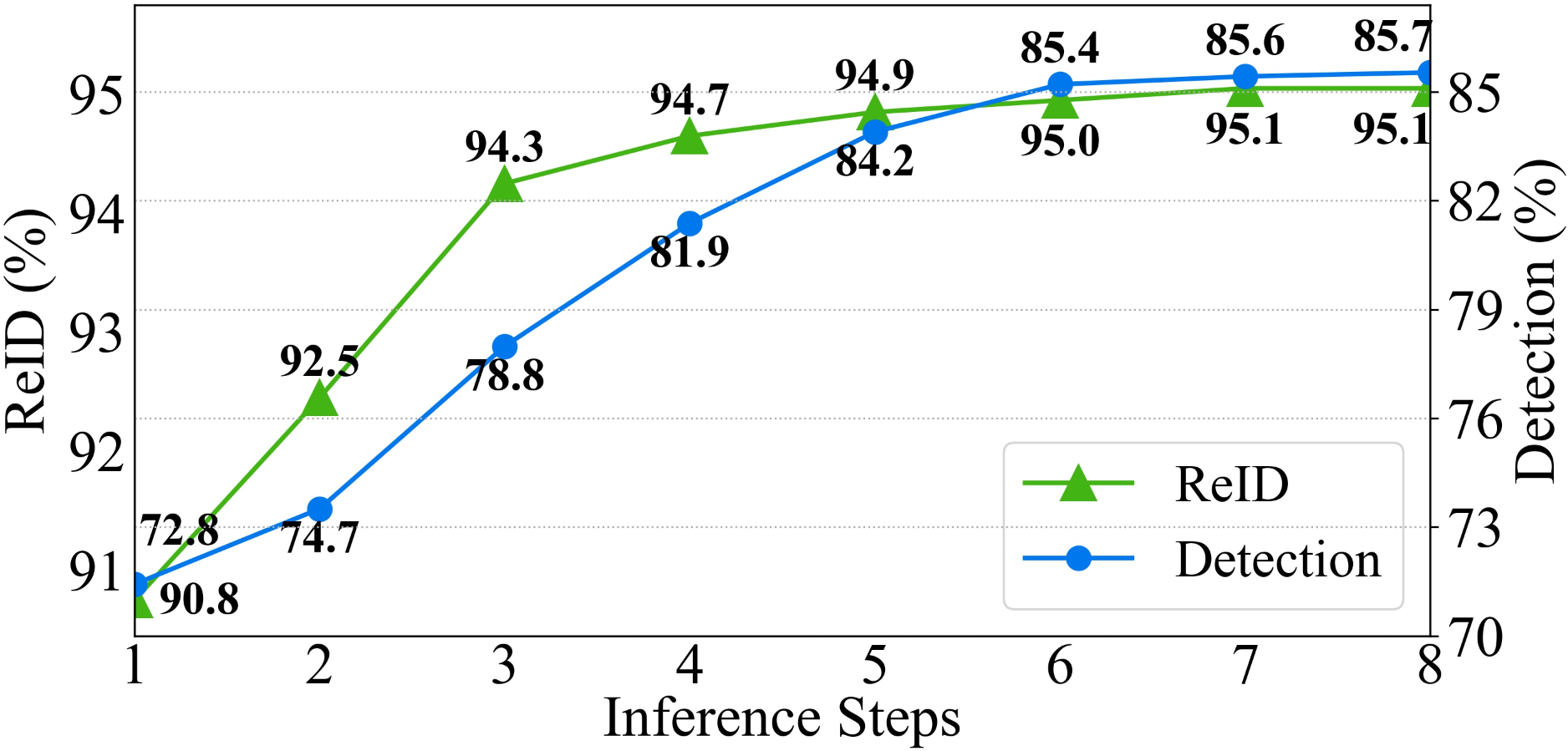}
     
   \caption{Performance with increasing steps.}
   \label{fig:steps}
\end{figure}

\begin{table}[t]
\renewcommand{\arraystretch}{1.2}
 	\tabcolsep=15 pt	
 	\begin{center}
        \caption{Results of different state-of-the-art Re-ID models as the teacher model in the PSDiff. The results on the left represent the performance obtained by utilizing teacher models through traditional distillation methods, while the underlined results on the right show the performance achieved by integrating the teacher model into our diffusion-based architecture.}
 		\begin{tabular}{c|cc}
 			\toprule
 			Methods  & mAP & top-1  \\
 			\midrule	
                 S-ReID \cite{HaoLuo2019ASB} + \underline{PSDiff} & 92.1 / \underline{93.5} & 93.0 / \underline{94.1} \\
                 MGN \cite{GuanshuoWang2018LearningDF} + \underline{PSDiff}  & 92.7 / \underline{93.9} & 92.3 / \underline{93.6}  \\
                 AGW \cite{MangYe2020DeepLF} + \underline{PSDiff} & 91.9 / \underline{93.1} & 92.5 / \underline{93.7}\\
                 PPLR \cite{cho2022part} + \underline{PSDiff} & 93.2 / \underline{94.3} & 93.0 / \underline{93.9}\\
                 \rowcolor{color1}HHCL \cite{hu2021hard} + \underline{PSDiff}  & 94.0 / \textbf{\underline{95.1}} & 93.9 / \textbf{\underline{95.3}}  \\
 			\bottomrule
 		\end{tabular}	
        \label{tab:kd}        
        \end{center}
\end{table}

\begin{table}[!t]
\tabcolsep=10 pt	
\centering
\caption{Computational cost. Training efficiency is measured in GPU hours.}
\label{tab:efficiency}
\begin{tabular}{cc|cc}
\toprule
Step   & Training & Step   & Training \\
\midrule
Teacher Model & 11 GPU hours  &
PSDiff  & 42 GPU hours  \\
\bottomrule
\end{tabular}
\end{table}

\subsection{Ablation Study}
In this section, we conduct extensive experiments on CUHK-SYSU to study PSDiff in detail. First, we analyze the effect of signal-to-noise ratios. Second, we show the effectiveness of our proposed collaborative interaction and cascaded prediction in CDL. Third, we further explore the effect of varying inference steps and fast sampling methods.

\subsubsection{Signal-to-Noise Ratios} 
As proposed in \cite{chen2022generalist,chen2022diffusiondet}, SNR has a significant effect on performance, especially for different tasks. 
Therefore, we study the influence of SNR for person search in Table \ref{tab:ablation} (a). Results demonstrate that SNR of $s_1 = 2.0$ for detection and $s_2 = 3.0$ for ReID achieve the best performance. Unlike the task of dense prediction that employs low SNR, \eg, 0.1 for panoptic segmentation\cite{chen2022generalist},  higher SNR means the denoising task for person search requires an easier training objective.

\subsubsection{Collaborative Interaction} We investigate the effectiveness of collaborative interaction in CDL.  As shown in Table \ref{tab:ablation} (b), the self-attention moudle can enhance detection features by reasoning about their visual relations, improving 1.2\% and 1.5\% on mAP and top-1. What's more, both interactions of the dynamic convolution and the cross-attention module are verified to be essential. Finally, the collaborative interaction significantly boosts the performance by 3.8\% and 3.5\% on mAP and top-1, respectively. 

\subsubsection{Cascaded prediction} To evaluate the effectiveness of the proposed cascaded prediction, we also implement the comparison with  “Parallel” strategy. As shown in Table \ref{tab:ablation} (c), our method performs better than the parallel prediction, especially for multi-steps inference. The proposed cascaded prediction outperforms the parallel prediction by 1.6\% mAP and 1.6\% top-1 accuracy, which demonstrates the superiority of our cascaded prediction.

\subsubsection{Iterative Inference} We first compare different fast sampling strategies in Table \ref{tab:sampling}. Results show more advanced methods, such as DPM-solver and DPM-solver++, bring slight benefits, indicating that the diffusion process for person search is insensitive with fast sampling methods. Therefore, for the sake of brevity, we default to employing DDIM as the fast sampling strategy. Then, we evaluate PSDiff with different inference steps in Fig. \ref{fig:steps}.  Thanks to our iterative and collaborative refinement, the performance of both detection and ReID consistently improves during the multi-step iterative process, without one task compromising the other, especially for detection performance from 72.8\% to 85.7\% on mAP50. This confirms that our method enables collaborative optimization between the two tasks, avoiding the pitfall of sub-optimization focused on a single task.
 We also note that the performance remains stable after 6 steps. It is very similar to DDIM for the generative task \cite{songdenoising}, where images generated with only 20 steps are slightly different from those generated with 1000 steps.

\subsection{Influence of Teacher Models in KD-based \emph{\textit{vs}} Diffusion-Based Approaches}
Table \ref{tab:kd} showcases the impact of employing different teacher models on person search. Overall, the performance fluctuates within a margin of 1-2\%. Notably, HHCL~\cite{hu2021hard} stands out with the best performance, owing to its effective use of hard samples and unlabeled samples. More importantly, when applying various methods as the teacher, our diffusion-based architecture consistently outperforms traditional distillation methods. This confirms that our approach minimizes performance loss.  What's more, the main increase in computational cost arises during training due to the incorporation of an additional teacher model. However, this component does not significantly extend the total training time, as detailed in Table \ref{tab:efficiency}, where we compare the computational overhead associated with each step. Importantly, the teacher model is not involved during inference, ensuring that our method does not incur extra computational costs at runtime.

\subsection{Qualitative Results}

Fig. \ref{fig:Case} visualizes the predicted detection and ReID results with increasing inference steps. As shown in cases in Fig. \ref{fig:Case}, each inference step gradually and collaboratively refines the position and confidence of predicted boxes and the matching score of predicted ReID embeddings. With our collaborative design, not only more accurate detection results contribute to learning more discriminative ReID embeddings, \eg, case (a) and case (c), but also more discriminative ReID clues help refine the position of predicted boxes and improve their confidence scores, \eg, case (b). After multiple refinements, final predictions are more accurate than primary predictions. 

\section{Conclusion}
\label{sec:conclusion}
In this work, we propose a novel diffusion-based person search framework, PSDiff, which formulates the task as a dual denoising process from noisy boxes and noisy ReID embeddings to ground truths. Following the new paradigm, the newly designed CDL enables us to reuse network parameters to refine predictions in an iterative and collaborative way, which also helps obtain the desired speed-accuracy trade-off. PSDiff successfully addresses the challenges of destructive pedestrian candidates and missing collaboration. Experiments on standard benchmarks show the PSDiff achieves SOTA performance with fewer parameters and elastic computing overhead. As a novel work in applying the diffusion model to the person search task, we aim to complement the scarce literature and provide a novel perspective for future research in this field.

\noindent \textbf{Limitations and Future Directions}: As a pioneering work exploring the use of diffusion models in person search, the proposed framework does have one primary limitation. In PSDiff, the ground truth for the diffusion process is derived from a teacher model, which serves as a temporary solution. This can introduce bias, as the teacher model is only a rough approximation of the true labels, limiting the upper bound of PSDiff’s performance. The quality of the ReID teacher model directly impacts the accuracy of the results. To address this, more flexible methods for representing ReID ground truths—such as a learnable embedding layer or tokenization techniques that map person IDs to ReID embeddings—could be explored. We discuss this limitation and anticipate that future research will build upon these. 




\ifCLASSOPTIONcaptionsoff
  \newpage
\fi

\bibliographystyle{IEEEtran}
\bibliography{mybib}

\begin{thebibliography}{100}
\providecommand{\url}[1]{#1}
\csname url@samestyle\endcsname
\providecommand{\newblock}{\relax}
\providecommand{\bibinfo}[2]{#2}
\providecommand{\BIBentrySTDinterwordspacing}{\spaceskip=0pt\relax}
\providecommand{\BIBentryALTinterwordstretchfactor}{4}
\providecommand{\BIBentryALTinterwordspacing}{\spaceskip=\fontdimen2\font plus
\BIBentryALTinterwordstretchfactor\fontdimen3\font minus
  \fontdimen4\font\relax}
\providecommand{\BIBforeignlanguage}[2]{{%
\expandafter\ifx\csname l@#1\endcsname\relax
\typeout{** WARNING: IEEEtran.bst: No hyphenation pattern has been}%
\typeout{** loaded for the language `#1'. Using the pattern for}%
\typeout{** the default language instead.}%
\else
\language=\csname l@#1\endcsname
\fi
#2}}
\providecommand{\BIBdecl}{\relax}
\BIBdecl

\bibitem{xiaoli2017joint}
T.~Xiao, S.~Li, B.~Wang, L.~Lin, and X.~Wang, ``Joint detection and
  identification feature learning for person search,'' in \emph{IEEE Conf.
  Comput. Vis. Pattern Recog.}, 2017.

\bibitem{Zheng_2017_CVPR}
L.~Zheng, H.~Zhang, S.~Sun, M.~Chandraker, Y.~Yang, and Q.~Tian, ``Person
  re-identification in the wild,'' in \emph{IEEE Conf. Comput. Vis. Pattern
  Recog.}, July 2017.

\bibitem{7918589}
H.~Liu, J.~Feng, M.~Qi, J.~Jiang, and S.~Yan, ``End-to-end comparative
  attention networks for person re-identification,'' \emph{IEEE Transactions on
  Image Processing}, vol.~26, no.~7, pp. 3492--3506, 2017.

\bibitem{8693885}
L.~Zheng, Y.~Huang, H.~Lu, and Y.~Yang, ``Pose-invariant embedding for deep
  person re-identification,'' \emph{IEEE Transactions on Image Processing},
  vol.~28, no.~9, pp. 4500--4509, 2019.

\bibitem{8485427}
Z.~Zhong, L.~Zheng, Z.~Zheng, S.~Li, and Y.~Yang, ``Camstyle: A novel data
  augmentation method for person re-identification,'' \emph{IEEE Transactions
  on Image Processing}, vol.~28, no.~3, pp. 1176--1190, 2019.

\bibitem{7864367}
A.~Wu, W.-S. Zheng, and J.-H. Lai, ``Robust depth-based person
  re-identification,'' \emph{IEEE Transactions on Image Processing}, vol.~26,
  no.~6, pp. 2588--2603, 2017.

\bibitem{9003518}
D.~Chen, S.~Zhang, W.~Ouyang, J.~Yang, and Y.~Tai, ``Person search by separated
  modeling and a mask-guided two-stream cnn model,'' \emph{IEEE Transactions on
  Image Processing}, vol.~29, pp. 4669--4682, 2020.

\bibitem{yan2019learning}
Y.~Yan, Q.~Zhang, B.~Ni, W.~Zhang, M.~Xu, and X.~Yang, ``Learning context graph
  for person search,'' in \emph{IEEE Conf. Comput. Vis. Pattern Recog.}, 2019,
  pp. 2158--2167.

\bibitem{chen2020norm}
D.~Chen, S.~Zhang, J.~Yang, and B.~Schiele, ``Norm-aware embedding for
  efficient person search,'' in \emph{IEEE Conf. Comput. Vis. Pattern Recog.},
  2020.

\bibitem{li2021sequential}
Z.~Li and D.~Miao, ``Sequential end-to-end network for efficient person
  search,'' in \emph{AAAI}, vol.~35, no.~3, 2021, pp. 2011--2019.

\bibitem{DBLP:conf/cvpr/YanLQBL00021}
Y.~Yan, J.~Li, J.~Qin, S.~Bai, S.~Liao, L.~Liu, F.~Zhu, and L.~Shao,
  ``Anchor-free person search,'' in \emph{IEEE Conf. Comput. Vis. Pattern
  Recog.}, 2021, pp. 7690--7699.

\bibitem{dong2020instance}
W.~Dong, Z.~Zhang, C.~Song, and T.~Tan, ``Instance guided proposal network for
  person search,'' in \emph{CVPR}, 2020, pp. 2585--2594.

\bibitem{Cao_PSTR_CVPR_2022}
J.~Cao, Y.~Pang, R.~M. Anwer, H.~Cholakkal, J.~Xie, M.~Shah, and F.~S. Khan,
  ``Pstr: End-to-end one-step person search with transformers,'' \emph{Proc.
  IEEE Conference on Computer Vision and Pattern Recognition}, 2022.

\bibitem{yu2022coat}
R.~Yu, D.~Du, R.~LaLonde, D.~Davila, C.~Funk, A.~Hoogs, and B.~Clipp, ``Cascade
  transformers for end-to-end person search,'' in \emph{{IEEE} Conference on
  Computer Vision and Pattern Recognition}, 2022.

\bibitem{rennips15fasterrcnn}
S.~Ren, K.~He, R.~Girshick, and J.~Sun, ``Faster {R-CNN}: Towards real-time
  object detection with region proposal networks,'' in \emph{Adv. Neural
  Inform. Process. Syst.}, 2015.

\bibitem{carion2020end}
N.~Carion, F.~Massa, G.~Synnaeve, N.~Usunier, A.~Kirillov, and S.~Zagoruyko,
  ``End-to-end object detection with transformers,'' in \emph{Computer
  Vision--ECCV 2020: 16th European Conference, Glasgow, UK, August 23--28,
  2020, Proceedings, Part I 16}.\hskip 1em plus 0.5em minus 0.4em\relax
  Springer, 2020, pp. 213--229.

\bibitem{ho2020denoising}
J.~Ho, A.~Jain, and P.~Abbeel, ``Denoising diffusion probabilistic models,''
  \emph{Advances in Neural Information Processing Systems}, vol.~33, pp.
  6840--6851, 2020.

\bibitem{Brempong_2022_CVPR}
E.~A. Brempong, S.~Kornblith, T.~Chen, N.~Parmar, M.~Minderer, and M.~Norouzi,
  ``Denoising pretraining for semantic segmentation,'' in \emph{Proceedings of
  the IEEE/CVF Conference on Computer Vision and Pattern Recognition (CVPR)
  Workshops}, June 2022, pp. 4175--4186.

\bibitem{chen2022diffusiondet}
S.~Chen, P.~Sun, Y.~Song, and P.~Luo, ``Diffusiondet: Diffusion model for
  object detection,'' in \emph{Proceedings of the IEEE/CVF international
  conference on computer vision}, 2023, pp. 19\,830--19\,843.

\bibitem{ye2021deep}
M.~Ye, J.~Shen, G.~Lin, T.~Xiang, L.~Shao, and S.~C. Hoi, ``Deep learning for
  person re-identification: A survey and outlook,'' \emph{IEEE transactions on
  pattern analysis and machine intelligence}, vol.~44, no.~6, pp. 2872--2893,
  2021.

\bibitem{wang2020unsupervised}
D.~Wang and S.~Zhang, ``Unsupervised person re-identification via multi-label
  classification,'' in \emph{Proceedings of the IEEE/CVF conference on computer
  vision and pattern recognition}, 2020, pp. 10\,981--10\,990.

\bibitem{9969623}
X.~Han, X.~Yu, G.~Li, J.~Zhao, G.~Pan, Q.~Ye, J.~Jiao, and Z.~Han, ``Rethinking
  sampling strategies for unsupervised person re-identification,'' \emph{IEEE
  Transactions on Image Processing}, vol.~32, pp. 29--42, 2023.

\bibitem{han2021group}
X.~Han, X.~Yu, N.~Jiang, G.~Li, J.~Zhao, Q.~Ye, and Z.~Han, ``Group sampling
  for unsupervised person re-identification,'' \emph{arXiv: 2107.03024}, 2021.

\bibitem{ge2020self}
Y.~Ge, F.~Zhu, D.~Chen, R.~Zhao \emph{et~al.}, ``Self-paced contrastive
  learning with hybrid memory for domain adaptive object re-id,''
  \emph{Advances in neural information processing systems}, vol.~33, pp.
  11\,309--11\,321, 2020.

\bibitem{zhao2020unsupervised}
F.~Zhao, S.~Liao, G.-S. Xie, J.~Zhao, K.~Zhang, and L.~Shao, ``Unsupervised
  domain adaptation with noise resistible mutual-training for person
  re-identification,'' in \emph{Computer Vision--ECCV 2020: 16th European
  Conference, Glasgow, UK, August 23--28, 2020, Proceedings, Part XI 16}.\hskip
  1em plus 0.5em minus 0.4em\relax Springer, 2020, pp. 526--544.

\bibitem{gu2023color}
J.~Gu, H.~Luo, K.~Wang, W.~Jiang, Y.~You, and J.~Zhao, ``Color prompting for
  data-free continual unsupervised domain adaptive person re-identification,''
  \emph{arXiv preprint arXiv:2308.10716}, 2023.

\bibitem{mclaughlin2016recurrent}
N.~McLaughlin, J.~M. Del~Rincon, and P.~Miller, ``Recurrent convolutional
  network for video-based person re-identification,'' in \emph{Proceedings of
  the IEEE conference on computer vision and pattern recognition}, 2016, pp.
  1325--1334.

\bibitem{9416694}
Z.~Wang, L.~He, X.~Tu, J.~Zhao, X.~Gao, S.~Shen, and J.~Feng, ``Robust
  video-based person re-identification by hierarchical mining,'' \emph{IEEE
  Transactions on Circuits and Systems for Video Technology}, vol.~32, no.~12,
  pp. 8179--8191, 2022.

\bibitem{9540797}
F.~Gurkan, L.~Cerkezi, O.~Cirakman, and B.~Gunsel, ``Tdiot: Target-driven
  inference for deep video object tracking,'' \emph{IEEE Transactions on Image
  Processing}, vol.~30, pp. 7938--7951, 2021.

\bibitem{6898845}
T.~Li, H.~Chang, M.~Wang, B.~Ni, R.~Hong, and S.~Yan, ``Crowded scene analysis:
  A survey,'' \emph{IEEE Transactions on Circuits and Systems for Video
  Technology}, vol.~25, no.~3, pp. 367--386, 2015.

\bibitem{4543858}
B.~T. Morris and M.~M. Trivedi, ``A survey of vision-based trajectory learning
  and analysis for surveillance,'' \emph{IEEE Transactions on Circuits and
  Systems for Video Technology}, vol.~18, no.~8, pp. 1114--1127, 2008.

\bibitem{7208833}
H.~Cheng, L.~Yang, and Z.~Liu, ``Survey on 3d hand gesture recognition,''
  \emph{IEEE Transactions on Circuits and Systems for Video Technology},
  vol.~26, no.~9, pp. 1659--1673, 2016.

\bibitem{han2019re}
C.~Han, J.~Ye, Y.~Zhong, X.~Tan, C.~Zhang, C.~Gao, and N.~Sang, ``Re-id driven
  localization refinement for person search,'' in \emph{Int. Conf. Comput.
  Vis.}, 2019, pp. 9814--9823.

\bibitem{wang2020tcts}
C.~Wang, B.~Ma, H.~Chang, S.~Shan, and X.~Chen, ``Tcts: A task-consistent
  two-stage framework for person search,'' in \emph{CVPR}, 2020, pp.
  11\,952--11\,961.

\bibitem{8640834}
Q.~Leng, M.~Ye, and Q.~Tian, ``A survey of open-world person
  re-identification,'' \emph{IEEE Transactions on Circuits and Systems for
  Video Technology}, vol.~30, no.~4, pp. 1092--1108, 2020.

\bibitem{8481710}
Z.~Zheng, L.~Zheng, and Y.~Yang, ``Pedestrian alignment network for large-scale
  person re-identification,'' \emph{IEEE Transactions on Circuits and Systems
  for Video Technology}, vol.~29, no.~10, pp. 3037--3045, 2019.

\bibitem{9285312}
X.~Ning, K.~Gong, W.~Li, L.~Zhang, X.~Bai, and S.~Tian, ``Feature refinement
  and filter network for person re-identification,'' \emph{IEEE Transactions on
  Circuits and Systems for Video Technology}, vol.~31, no.~9, pp. 3391--3402,
  2021.

\bibitem{6490028}
D.~Tao, L.~Jin, Y.~Wang, Y.~Yuan, and X.~Li, ``Person re-identification by
  regularized smoothing kiss metric learning,'' \emph{IEEE Transactions on
  Circuits and Systems for Video Technology}, vol.~23, no.~10, pp. 1675--1685,
  2013.

\bibitem{7948787}
H.~Liu, Z.~Jie, K.~Jayashree, M.~Qi, J.~Jiang, S.~Yan, and J.~Feng,
  ``Video-based person re-identification with accumulative motion context,''
  \emph{IEEE Transactions on Circuits and Systems for Video Technology},
  vol.~28, no.~10, pp. 2788--2802, 2018.

\bibitem{7979595}
D.~Tao, Y.~Guo, B.~Yu, J.~Pang, and Z.~Yu, ``Deep multi-view feature learning
  for person re-identification,'' \emph{IEEE Transactions on Circuits and
  Systems for Video Technology}, vol.~28, no.~10, pp. 2657--2666, 2018.

\bibitem{7279146}
C.~Zhu and Y.~Peng, ``A boosted multi-task model for pedestrian detection with
  occlusion handling,'' \emph{IEEE Transactions on Image Processing}, vol.~24,
  no.~12, pp. 5619--5629, 2015.

\bibitem{8328854}
X.~Zhang, L.~Cheng, B.~Li, and H.-M. Hu, ``Too far to see? not
  really!—pedestrian detection with scale-aware localization policy,''
  \emph{IEEE Transactions on Image Processing}, vol.~27, no.~8, pp. 3703--3715,
  2018.

\bibitem{9282190}
J.~Xie, Y.~Pang, M.~H. Khan, R.~M. Anwer, F.~S. Khan, and L.~Shao,
  ``Mask-guided attention network and occlusion-sensitive hard example mining
  for occluded pedestrian detection,'' \emph{IEEE Transactions on Image
  Processing}, vol.~30, pp. 3872--3884, 2021.

\bibitem{song2021scorebased}
\BIBentryALTinterwordspacing
Y.~Song, J.~Sohl-Dickstein, D.~P. Kingma, A.~Kumar, S.~Ermon, and B.~Poole,
  ``Score-based generative modeling through stochastic differential
  equations,'' in \emph{International Conference on Learning Representations},
  2021. [Online]. Available: \url{https://openreview.net/forum?id=PxTIG12RRHS}
\BIBentrySTDinterwordspacing

\bibitem{dhariwal2021diffusion}
P.~Dhariwal and A.~Nichol, ``Diffusion models beat gans on image synthesis,''
  \emph{Advances in Neural Information Processing Systems}, vol.~34, pp.
  8780--8794, 2021.

\bibitem{rombach2022high}
R.~Rombach, A.~Blattmann, D.~Lorenz, P.~Esser, and B.~Ommer, ``High-resolution
  image synthesis with latent diffusion models,'' in \emph{Proceedings of the
  IEEE/CVF Conference on Computer Vision and Pattern Recognition}, 2022, pp.
  10\,684--10\,695.

\bibitem{jia2024ssmg}
C.~Jia, M.~Luo, Z.~Dang, G.~Dai, X.~Chang, M.~Wang, and J.~Wang, ``Ssmg:
  Spatial-semantic map guided diffusion model for free-form layout-to-image
  generation,'' in \emph{Proceedings of the AAAI Conference on Artificial
  Intelligence}, vol.~38, no.~3, 2024, pp. 2480--2488.

\bibitem{nichol2021glide}
A.~Q. Nichol, P.~Dhariwal, A.~Ramesh, P.~Shyam, P.~Mishkin, B.~Mcgrew,
  I.~Sutskever, and M.~Chen, ``{GLIDE}: Towards photorealistic image generation
  and editing with text-guided diffusion models,'' in \emph{Proceedings of the
  39th International Conference on Machine Learning}, ser. Proceedings of
  Machine Learning Research, K.~Chaudhuri, S.~Jegelka, L.~Song, C.~Szepesvari,
  G.~Niu, and S.~Sabato, Eds., vol. 162.\hskip 1em plus 0.5em minus 0.4em\relax
  PMLR, 17--23 Jul 2022, pp. 16\,784--16\,804.

\bibitem{saharia2022image}
C.~Saharia, J.~Ho, W.~Chan, T.~Salimans, D.~J. Fleet, and M.~Norouzi, ``Image
  super-resolution via iterative refinement,'' \emph{IEEE Transactions on
  Pattern Analysis and Machine Intelligence}, 2022.

\bibitem{amit2021segdiff}
T.~Amit, E.~Nachmani, T.~Shaharbany, and L.~Wolf, ``Segdiff: Image segmentation
  with diffusion probabilistic models,'' \emph{arXiv preprint
  arXiv:2112.00390}, 2021.

\bibitem{chen2022generalist}
T.~Chen, L.~Li, S.~Saxena, G.~Hinton, and D.~J. Fleet, ``A generalist framework
  for panoptic segmentation of images and videos,'' in \emph{Proceedings of the
  IEEE/CVF international conference on computer vision}, 2023, pp. 909--919.

\bibitem{han2022card}
X.~Han, H.~Zheng, and M.~Zhou, ``Card: Classification and regression diffusion
  models,'' in \emph{Thirty-Sixth Conference on Neural Information Processing
  Systems}, 2022.

\bibitem{pinaya2022fast}
W.~H. Pinaya, M.~S. Graham, R.~Gray, P.~F. Da~Costa, P.-D. Tudosiu, P.~Wright,
  Y.~H. Mah, A.~D. MacKinnon, J.~T. Teo, R.~Jager \emph{et~al.}, ``Fast
  unsupervised brain anomaly detection and segmentation with diffusion
  models,'' in \emph{Medical Image Computing and Computer Assisted
  Intervention--MICCAI 2022: 25th International Conference, Singapore,
  September 18--22, 2022, Proceedings, Part VIII}.\hskip 1em plus 0.5em minus
  0.4em\relax Springer, 2022, pp. 705--714.

\bibitem{wolleb2022diffusion}
J.~Wolleb, F.~Bieder, R.~Sandk{\"u}hler, and P.~C. Cattin, ``Diffusion models
  for medical anomaly detection,'' in \emph{Medical Image Computing and
  Computer Assisted Intervention--MICCAI 2022: 25th International Conference,
  Singapore, September 18--22, 2022, Proceedings, Part VIII}.\hskip 1em plus
  0.5em minus 0.4em\relax Springer, 2022, pp. 35--45.

\bibitem{9857019}
J.~Wyatt, A.~Leach, S.~M. Schmon, and C.~G. Willcocks, ``Anoddpm: Anomaly
  detection with denoising diffusion probabilistic models using simplex
  noise,'' in \emph{2022 IEEE/CVF Conference on Computer Vision and Pattern
  Recognition Workshops (CVPRW)}, 2022, pp. 649--655.

\bibitem{pandey2022diffusevae}
K.~Pandey, A.~Mukherjee, P.~Rai, and A.~Kumar, ``Diffusevae: Efficient,
  controllable and high-fidelity generation from low-dimensional latents,''
  \emph{arXiv preprint arXiv:2201.00308}, 2022.

\bibitem{ho2022cascaded}
J.~Ho, C.~Saharia, W.~Chan, D.~J. Fleet, M.~Norouzi, and T.~Salimans,
  ``Cascaded diffusion models for high fidelity image generation.'' \emph{J.
  Mach. Learn. Res.}, vol.~23, no.~47, pp. 1--33, 2022.

\bibitem{sohl2015deep}
J.~Sohl-Dickstein, E.~Weiss, N.~Maheswaranathan, and S.~Ganguli, ``Deep
  unsupervised learning using nonequilibrium thermodynamics,'' in
  \emph{International Conference on Machine Learning}.\hskip 1em plus 0.5em
  minus 0.4em\relax PMLR, 2015, pp. 2256--2265.

\bibitem{nichol2021improved}
A.~Q. Nichol and P.~Dhariwal, ``Improved denoising diffusion probabilistic
  models,'' in \emph{International Conference on Machine Learning}.\hskip 1em
  plus 0.5em minus 0.4em\relax PMLR, 2021, pp. 8162--8171.

\bibitem{songdenoising}
J.~Song, C.~Meng, and S.~Ermon, ``Denoising diffusion implicit models,'' in
  \emph{International Conference on Learning Representations}.

\bibitem{lin2017feature}
T.-Y. Lin, P.~Doll{\'a}r, R.~Girshick, K.~He, B.~Hariharan, and S.~Belongie,
  ``Feature pyramid networks for object detection,'' in \emph{Proceedings of
  the IEEE conference on computer vision and pattern recognition}, 2017, pp.
  2117--2125.

\bibitem{he2016deep}
K.~He, X.~Zhang, S.~Ren, and J.~Sun, ``Deep residual learning for image
  recognition,'' in \emph{Proceedings of the IEEE conference on computer vision
  and pattern recognition}, 2016, pp. 770--778.

\bibitem{liu2021swin}
Z.~Liu, Y.~Lin, Y.~Cao, H.~Hu, Y.~Wei, Z.~Zhang, S.~Lin, and B.~Guo, ``Swin
  transformer: Hierarchical vision transformer using shifted windows,'' in
  \emph{Proceedings of the IEEE/CVF international conference on computer
  vision}, 2021, pp. 10\,012--10\,022.

\bibitem{10234225}
C.~Jia, M.~Luo, C.~Yan, L.~Zhu, X.~Chang, and Q.~Zheng, ``Collaborative
  contrastive refining for weakly supervised person search,'' \emph{IEEE
  Transactions on Image Processing}, pp. 1--1, 2023.

\bibitem{hu2021hard}
Z.~Hu, C.~Zhu, and G.~He, ``Hard-sample guided hybrid contrast learning for
  unsupervised person re-identification,'' in \emph{Int. Conf. Comput. Vis.},
  2021.

\bibitem{luo2021dual}
Y.~Luo, J.~Ji, X.~Sun, L.~Cao, Y.~Wu, F.~Huang, C.-W. Lin, and R.~Ji,
  ``Dual-level collaborative transformer for image captioning,'' in
  \emph{Proceedings of the AAAI conference on artificial intelligence},
  vol.~35, no.~3, 2021, pp. 2286--2293.

\bibitem{sun2021sparse}
P.~Sun, R.~Zhang, Y.~Jiang, T.~Kong, C.~Xu, W.~Zhan, M.~Tomizuka, L.~Li,
  Z.~Yuan, C.~Wang \emph{et~al.}, ``Sparse r-cnn: End-to-end object detection
  with learnable proposals,'' in \emph{Proceedings of the IEEE/CVF conference
  on computer vision and pattern recognition}, 2021, pp. 14\,454--14\,463.

\bibitem{cai2018cascade}
Z.~Cai and N.~Vasconcelos, ``Cascade r-cnn: Delving into high quality object
  detection,'' in \emph{Proceedings of the IEEE conference on computer vision
  and pattern recognition}, 2018, pp. 6154--6162.

\bibitem{stewart2016end}
R.~Stewart, M.~Andriluka, and A.~Y. Ng, ``End-to-end people detection in
  crowded scenes,'' in \emph{Proceedings of the IEEE conference on computer
  vision and pattern recognition}, 2016, pp. 2325--2333.

\bibitem{lin2017focal}
T.-Y. Lin, P.~Goyal, R.~Girshick, K.~He, and P.~Doll{\'a}r, ``Focal loss for
  dense object detection,'' in \emph{Proceedings of the IEEE international
  conference on computer vision}, 2017, pp. 2980--2988.

\bibitem{rezatofighi2019generalized}
H.~Rezatofighi, N.~Tsoi, J.~Gwak, A.~Sadeghian, I.~Reid, and S.~Savarese,
  ``Generalized intersection over union: A metric and a loss for bounding box
  regression,'' in \emph{Proceedings of the IEEE/CVF conference on computer
  vision and pattern recognition}, 2019, pp. 658--666.

\bibitem{ludpm}
C.~Lu, Y.~Zhou, F.~Bao, J.~Chen, C.~Li, and J.~Zhu, ``Dpm-solver: A fast ode
  solver for diffusion probabilistic model sampling in around 10 steps,'' in
  \emph{Advances in Neural Information Processing Systems}.

\bibitem{lu2022dpm}
------, ``Dpm-solver++: Fast solver for guided sampling of diffusion
  probabilistic models,'' \emph{arXiv preprint arXiv:2211.01095}, 2022.

\bibitem{han2021decoupled}
C.~Han, Z.~Zheng, C.~Gao, N.~Sang, and Y.~Yang, ``Decoupled and
  memory-reinforced networks: Towards effective feature learning for one-step
  person search,'' in \emph{Proceedings of the AAAI Conference on Artificial
  Intelligence}, vol.~35, no.~2, 2021, pp. 1505--1512.

\bibitem{yang2023towards}
S.~Yang, Y.~Zhou, Z.~Zheng, Y.~Wang, L.~Zhu, and Y.~Wu, ``Towards unified
  text-based person retrieval: A large-scale multi-attribute and language
  search benchmark,'' in \emph{Proceedings of the 31st ACM International
  Conference on Multimedia}, 2023, pp. 4492--4501.

\bibitem{he2024instruct}
W.~He, Y.~Deng, S.~Tang, Q.~Chen, Q.~Xie, Y.~Wang, L.~Bai, F.~Zhu, R.~Zhao,
  W.~Ouyang \emph{et~al.}, ``Instruct-reid: A multi-purpose person
  re-identification task with instructions,'' in \emph{Proceedings of the
  IEEE/CVF Conference on Computer Vision and Pattern Recognition}, 2024, pp.
  17\,521--17\,531.

\bibitem{glorot2010understanding}
X.~Glorot and Y.~Bengio, ``Understanding the difficulty of training deep
  feedforward neural networks,'' in \emph{Proceedings of the thirteenth
  international conference on artificial intelligence and statistics}.\hskip
  1em plus 0.5em minus 0.4em\relax JMLR Workshop and Conference Proceedings,
  2010, pp. 249--256.

\bibitem{loshchilov2017decoupled}
\BIBentryALTinterwordspacing
I.~Loshchilov and F.~Hutter, ``Decoupled weight decay regularization,'' in
  \emph{International Conference on Learning Representations}, 2019. [Online].
  Available: \url{https://openreview.net/forum?id=Bkg6RiCqY7}
\BIBentrySTDinterwordspacing

\bibitem{chang2018rcaa}
X.~Chang, P.-Y. Huang, Y.-D. Shen, X.~Liang, Y.~Yang, and A.~G. Hauptmann,
  ``Rcaa: Relational context-aware agents for person search,'' in \emph{Eur.
  Conf. Comput. Vis.}, 2018, pp. 84--100.

\bibitem{9265450}
H.~Yao and C.~Xu, ``Joint person objectness and repulsion for person search,''
  \emph{IEEE Transactions on Image Processing}, vol.~30, pp. 685--696, 2021.

\bibitem{Yan_2019_CVPR}
Y.~Yan, Q.~Zhang, B.~Ni, W.~Zhang, M.~Xu, and X.~Yang, ``Learning context graph
  for person search,'' in \emph{IEEE Conf. Comput. Vis. Pattern Recog.}, June
  2019.

\bibitem{munjal2019knowledge}
B.~Munjal, S.~Amin, and F.~Galasso, ``Knowledge distillation for end-to-end
  person search,'' in \emph{The British Machine Vision Conference (BMVC)},
  2019.

\bibitem{chen2020hoim}
D.~Chen, S.~Zhang, W.~Ouyang, J.~Yang, and B.~Schiele, ``Hierarchical online
  instance matching for person search,'' in \emph{AAAI}, 2020.

\bibitem{Dong_2020_CVPR}
W.~Dong, Z.~Zhang, C.~Song, and T.~Tan, ``Bi-directional interaction network
  for person search,'' in \emph{IEEE Conf. Comput. Vis. Pattern Recog.}, June
  2020.

\bibitem{8759990}
Y.~Zhang, X.~Li, and Z.~Zhang, ``Efficient person search via expert-guided
  knowledge distillation,'' \emph{IEEE Transactions on Cybernetics}, vol.~51,
  no.~10, pp. 5093--5104, 2021.

\bibitem{zhang2021diverse}
X.~Zhang, X.~Wang, J.-W. Bian, C.~Shen, and M.~You, ``Diverse knowledge
  distillation for end-to-end person search,'' in \emph{Proc. AAAI Conf. Artif.
  Intell.}, vol.~35, no.~4, 2021, pp. 3412--3420.

\bibitem{Kim_2021_CVPR}
H.~Kim, S.~Joung, I.-J. Kim, and K.~Sohn, ``Prototype-guided saliency feature
  learning for person search,'' in \emph{Proceedings of the IEEE/CVF Conference
  on Computer Vision and Pattern Recognition (CVPR)}, June 2021, pp.
  4865--4874.

\bibitem{Han_2021_ICCV}
B.-J. Han, K.~Ko, and J.-Y. Sim, ``End-to-end trainable trident person search
  network using adaptive gradient propagation,'' in \emph{Int. Conf. Comput.
  Vis.}, October 2021, pp. 925--933.

\bibitem{9352705}
W.~Yang, H.~Huang, X.~Chen, and K.~Huang, ``Bottom-up foreground-aware feature
  fusion for practical person search,'' \emph{IEEE Transactions on Circuits and
  Systems for Video Technology}, vol.~32, no.~1, pp. 262--274, 2022.

\bibitem{9438639}
S.~Hou, C.~Zhao, Z.~Chen, J.~Wu, Z.~Wei, and D.~Miao, ``Improved instance
  discrimination and feature compactness for end-to-end person search,''
  \emph{IEEE Transactions on Circuits and Systems for Video Technology},
  vol.~32, no.~4, pp. 2079--2090, 2022.

\bibitem{lee2022oimnet++}
S.~Lee, Y.~Oh, D.~Baek, J.~Lee, and B.~Ham, ``Oimnet++: Prototypical
  normalization and localization-aware learning for person search,'' in
  \emph{European Conference on Computer Vision}.\hskip 1em plus 0.5em minus
  0.4em\relax Springer, 2022, pp. 621--637.

\bibitem{9785793}
C.~Zhao, Z.~Chen, S.~Dou, Z.~Qu, J.~Yao, J.~Wu, and D.~Miao, ``Context-aware
  feature learning for noise robust person search,'' \emph{IEEE Transactions on
  Circuits and Systems for Video Technology}, vol.~32, no.~10, pp. 7047--7060,
  2022.

\bibitem{10095679}
C.~Jia, M.~Luo, Z.~Dang, X.~Chang, and Q.~Zheng, ``Towards real-time person
  search with invariant feature learning,'' in \emph{ICASSP 2023 - 2023 IEEE
  International Conference on Acoustics, Speech and Signal Processing
  (ICASSP)}, 2023, pp. 1--5.

\bibitem{ZHANG2024110053}
\BIBentryALTinterwordspacing
P.~Zhang, X.~Yu, X.~Bai, C.~Wang, J.~Zheng, and X.~Ning, ``Joint discriminative
  representation learning for end-to-end person search,'' \emph{Pattern
  Recognition}, vol. 147, p. 110053, 2024. [Online]. Available:
  \url{https://www.sciencedirect.com/science/article/pii/S0031320323007501}
\BIBentrySTDinterwordspacing

\bibitem{jaffe2023gallery}
L.~Jaffe and A.~Zakhor, ``Gallery filter network for person search,'' in
  \emph{Proceedings of the IEEE/CVF Winter Conference on Applications of
  Computer Vision}, 2023, pp. 1684--1693.

\bibitem{10599306}
Z.~Li, Y.~Shi, H.~Ling, J.~Chen, R.~Wang, C.~Zhao, Q.~Wang, and S.~Huang,
  ``Knowledge consistency distillation for weakly supervised one step person
  search,'' \emph{IEEE Transactions on Circuits and Systems for Video
  Technology}, pp. 1--1, 2024.

\bibitem{tian2023divide}
Y.~Tian, D.~Chen, Y.~Liu, J.~Yang, and S.~Zhang, ``Divide and conquer: Hybrid
  pre-training for person search,'' \emph{AAAI}, 2024.

\bibitem{li2022domain}
J.~Li, Y.~Yan, G.~Wang, F.~Yu, Q.~Jia, and S.~Ding, ``Domain adaptive person
  search,'' in \emph{European Conference on Computer Vision}.\hskip 1em plus
  0.5em minus 0.4em\relax Springer, 2022, pp. 302--318.

\bibitem{HaoLuo2019ASB}
H.~Luo, W.~Jiang, Y.~Gu, F.~Liu, X.~Liao, S.~Lai, and J.~Gu, ``A strong
  baseline and batch normalization neck for deep person re-identification,''
  2019.

\bibitem{GuanshuoWang2018LearningDF}
G.~Wang, Y.~Yuan, X.~Chen, J.~Li, and X.~Zhou, ``Learning discriminative
  features with multiple granularities for person re-identification,'' 2018.

\bibitem{MangYe2020DeepLF}
M.~Ye, J.~Shen, G.~Lin, T.~Xiang, L.~Shao, and S.~C.~H. Hoi, ``Deep learning
  for person re-identification: A survey and outlook,'' 2020.

\bibitem{cho2022part}
Y.~Cho, W.~J. Kim, S.~Hong, and S.-E. Yoon, ``Part-based pseudo label
  refinement for unsupervised person re-identification,'' in \emph{Proceedings
  of the IEEE/CVF Conference on Computer Vision and Pattern Recognition}, 2022,
  pp. 7308--7318.

\end{thebibliography}

 

\begin{IEEEbiography}[{\includegraphics[width=1in,height=1.25in,clip,keepaspectratio]{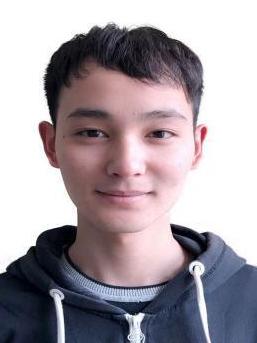}}]{Chengyou Jia} received the BS degree in Computer Science and Technology from Xi’an Jiaotong University in 2021. He is currently working toward the Ph. D. degree in Computer Science and Technology at Xi’an Jiaotong University. His
research interests include machine learning and
optimization, computer vision and multi-modal learning.
\end{IEEEbiography}

\begin{IEEEbiography}[{\includegraphics[width=1in,height=1.25in,clip,keepaspectratio]{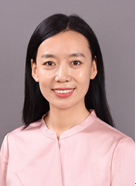}}]{Minnan Luo} received the Ph. D. degree from
the Department of Computer Science and Technology, Tsinghua University, China, in 2014.
Currently, she is a Professor in the
School of Electronic and Information Engineering at Xi’an Jiaotong University. She was a PostDoctoral Research with the School of Computer
Science, Carnegie Mellon University, Pittsburgh,
PA, USA. Her research interests include machine learning and optimization, cross-media retrieval and fuzzy system.

\end{IEEEbiography}

\begin{IEEEbiography}[{\includegraphics[width=1in,height=1.25in,clip,keepaspectratio]{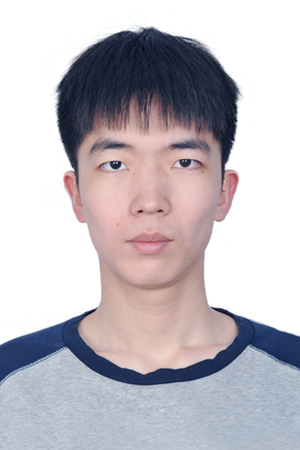}}]{Zhuohang Dang} received a BS degree from the Department of Computer Science and Technology, Xi'an Jiaotong University, in 2021. He is currently working toward the Ph. D. degree in Computer Science and Technology at Xi’an Jiaotong University, supervised by Prof. Minnan Luo. His research interests include causal inference and computer vision.
\end{IEEEbiography}

\begin{IEEEbiography}[{\includegraphics[width=1in,height=1.25in,clip,keepaspectratio]{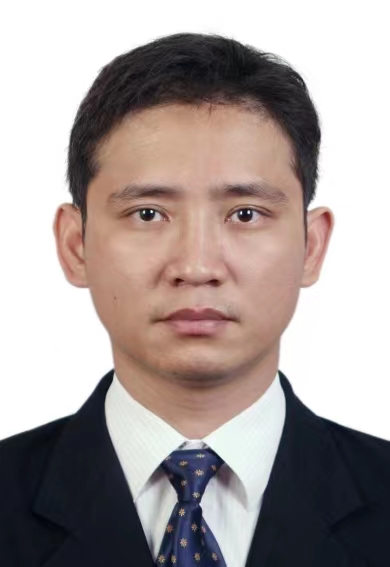}}]{Guang Dai} received his B.Eng. degree in Mechanical Engineering from Dalian University of Technology and M.Phil. degree in Computer Science from the Zhejiang University and the Hong Kong University of Science and Technology. He is currently a senior research scientist at State Grid Corporation of China. He has published a number of papers at prestigious journals and conferences, e.g., JMLR, AIJ, PR, NeurIPS, ICML, AISTATS, IJCAI, AAAI, ECML, CVPR, ECCV. His main research interests include Bayesian statistics, deep learning, reinforcement learning, and related applications.
\end{IEEEbiography}

\begin{IEEEbiography}[{\includegraphics[width=1in,height=1.25in,clip,keepaspectratio]{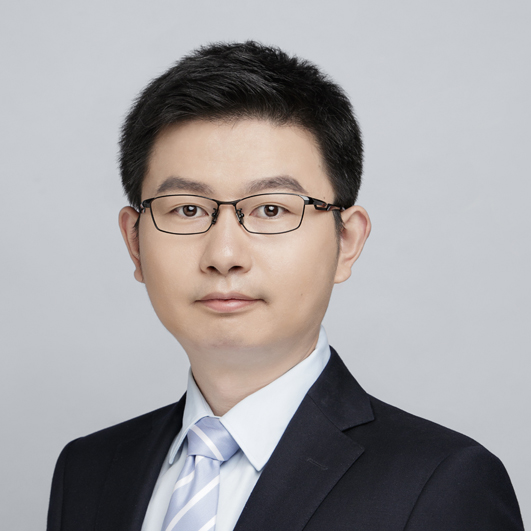}}]{Xiaojun Chang} is a Professor at the School of Information Science and Technology, University of Science and Technology of China. He is also a visiting Professor at Department of Computer Vision, Mohamed bin Zayed University of Artificial Intelligence (MBZUAI). He was an ARC Discovery Early Career Researcher Award (DECRA)
Fellow between 2019-2021. After graduation, he
was worked as a Postdoc Research Associate
in School of Computer Science, Carnegie Mellon University, a Senior Lecturer in Faculty of
Information Technology, Monash University, an Associate Professor in School of Computing Technologies, RMIT University, and a Professor in Faculty of Engineering and Information Technology, University of Technology Sydney. He mainly
worked on exploring multiple signals for automatic content analysis in unconstrained or surveillance videos and has achieved top performance in various international competitions. He received his Ph.D. degree from University of Technology Sydney. His research focus in this period was mainly on developing machine learning algorithms and applying them to
multimedia analysis and computer vision.
\end{IEEEbiography}

\begin{IEEEbiography}[{\includegraphics[width=1in,height=1.25in,clip,keepaspectratio]{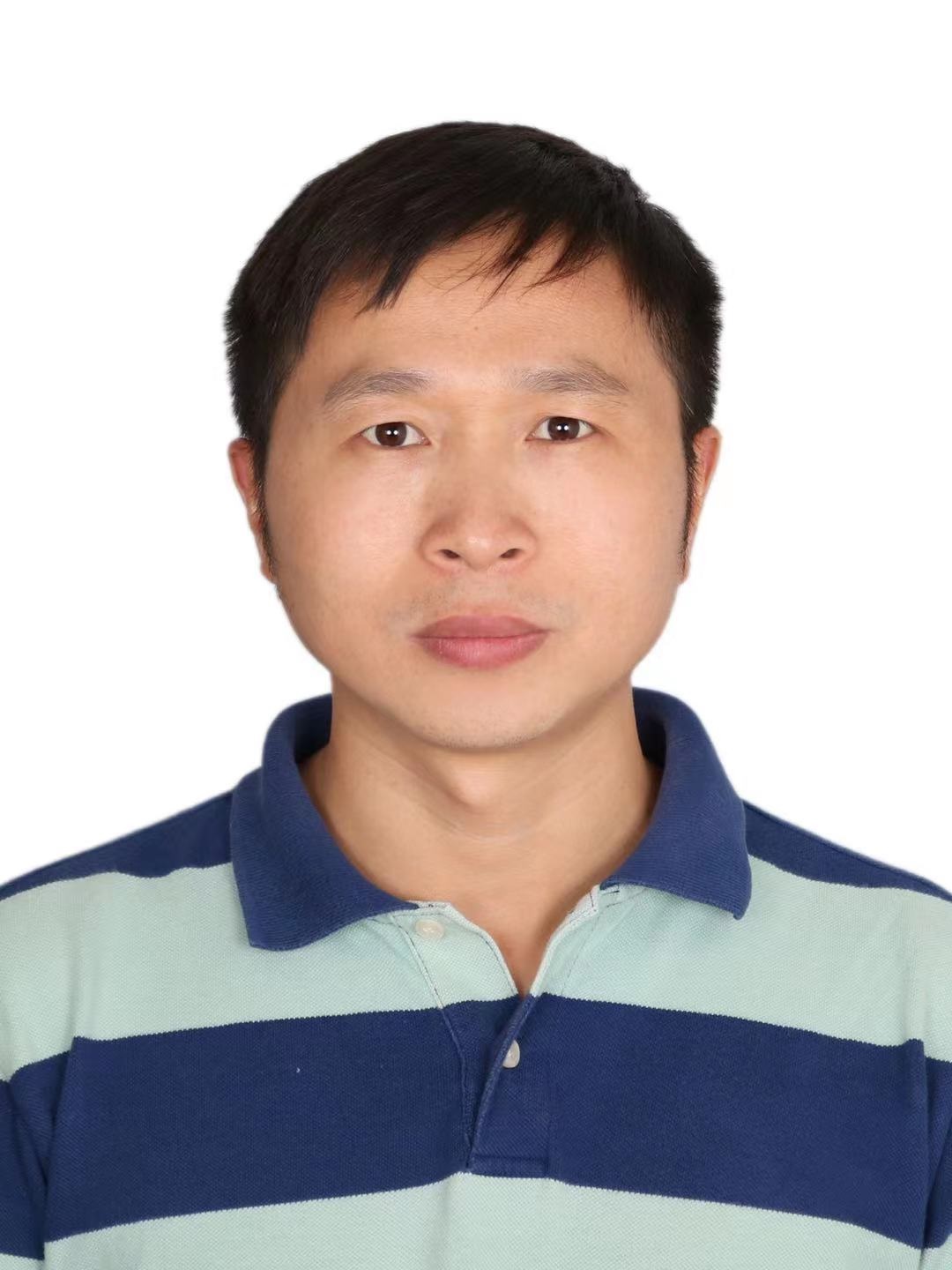}}]{Jingdong Wang} is Chief Scientist for computer vision with Baidu. Before joining Baidu, he was a Senior Principal Researcher at Microsoft Research Asia from September 2007 to August 2021. His areas of interest include vision foundation models, self-supervised pretraining, OCR, human pose estimation, semantic segmentation, image classification, object detection, and large-scale indexing. His representative works include high-resolution network (HRNet) for generic visual recognition, object-contextual representations (OCRNet) for semantic segmentation discriminative regional feature integration (DRFI) for saliency detection, neighborhood graph search (NGS, SPTAG) for vector search. He has been serving/served as an Associate Editor of IEEE TPAMI, IJCV, IEEE TMM, and IEEE TCSVT, and an (senior) area chair of leading conferences in vision, multimedia, and AI, such as CVPR, ICCV, ECCV, ACM MM, IJCAI, and AAAI. He was elected as an ACM Distinguished Member, a Fellow of IAPR, and a Fellow of IEEE, for his contributions to visual content understanding and retrieval.
\end{IEEEbiography}

\vfill

\end{document}